\documentclass{article}

\usepackage[nonatbib, preprint]{neurips_2022}

\usepackage[
    natbib=true,
    style=numeric,
    sorting=none
]{biblatex}
\usepackage[utf8]{inputenc} % allow utf-8 input
\usepackage[T1]{fontenc}    % use 8-bit T1 fonts
\usepackage{hyperref}       % hyperlinks
\usepackage{url}            % simple URL typesetting
\usepackage{booktabs}       % professional-quality tables
\usepackage{amsfonts}       % blackboard math symbols
\usepackage{nicefrac}       % compact symbols for 1/2, etc.
\usepackage{microtype}      % microtypography
\usepackage{xcolor}         % colors
\usepackage{float}          % suppressing table floating
\usepackage{graphicx}       % images
\usepackage{siunitx}        % scientific units
\usepackage{multirow}       % multirow support
\usepackage{amsopn}         % operatorname support

\DeclareSIUnit\angstrom{\text {Å}}

\title{\textsc{DProQ}: A Gated-Graph Transformer for Protein Complex Structure Assessment}

\author{
  Xiao Chen\thanks{Equal contribution.}, Alex Morehead\thanks{Equal contribution.}, Jian Liu, Jianlin Cheng\\
  Department of Electrical Engineering \& Computer Science \\
  University of Missouri \\
  Columbia, MO, USA \\
  \texttt{\{xcbh6, acmwhb, jl4mc, chengji\}@umsystem.edu}\\
}

\begin{document}

\maketitle

\begin{abstract}
    Proteins interact to form complexes to carry out essential biological functions. Computational methods have been developed to predict the structures of protein complexes. However, an important challenge in protein complex structure prediction is to estimate the quality of predicted protein complex structures without any knowledge of the corresponding native structures. Such estimations can then be used to select high-quality predicted complex structures to facilitate biomedical research such as protein function analysis and drug discovery. We challenge this significant task with \textsc{DProQ}, which introduces a gated neighborhood-modulating Graph Transformer (\textsc{GGT}) designed to predict the quality of 3D protein complex structures. Notably, we incorporate node and edge gates within a novel Graph Transformer framework to control information flow during graph message passing. We train and evaluate \textsc{DProQ} on four newly-developed datasets that we make publicly available in this work. Our rigorous experiments demonstrate that \textsc{DProQ} achieves state-of-the-art performance in ranking protein complex structures.\footnote{Source code, data, and pre-trained models are available at \href{https://github.com/BioinfoMachineLearning/DProQ}{\texttt{https://github.com/BioinfoMachineLearning/DProQ}}}
\end{abstract}

\section{Introduction}
Proteins, bio-molecules, or biological macro-molecules perform a broad range of functions in modern biology and bioinformatics. Protein-protein interactions (PPI) play a key role in almost all biological processes. Understanding the mechanisms and functions of PPI may benefit studies in other scientific areas such as drug discovery \cite{scott2016small, athanasios2017protein, macalino2018evolution} and protein design \cite{kortemme2004computational, baker2006prediction, lippow2007progress}. Typically, high-resolution 3D structures of protein complexes can be determined using experimental solutions (e.g., X-ray crystallography and cryo-electron microscopy). However, due to the high financial and resource-intensive costs associated with them, these methods are not satisfactory for the increasing demands in modern biology research. In the context of this practical challenge, computational methods for protein complex structure prediction have recently been receiving an increasing amount of attention.

Recently, the \textit{ab initio} method AlphaFold-Multimer \cite{evans2021protein} released an end-to-end system for protein complex structure prediction. The system improves its prediction accuracy on multimeric proteins considerably. However, compared to AlphaFold2's outstanding performance in monomer structure prediction \cite{jumper2021highly}, the accuracy level for protein quaternary structure prediction still has much room for progress. Within this context, estimation of model accuracy methods (EMA) and quality assessment methods (QA) play a significant role in the advancement of protein complex structure prediction \cite{kinch2021topology}.

However, existing EMA methods have two primary shortcomings. The first is that they have not explored the use of Transformer-like architectures \cite{vaswani2017attention} to enhance the expressiveness of neural networks designed for structure quality assessment. The second shortcoming is that current training or benchmark datasets for QA \cite{liu2008dockground, lensink2014score_set, kotthoff2021dockground} cannot fully represent the accuracy level of the latest protein complex structure predictors \cite{xie2022deep, jumper2021highly, bryant2022improved}. Most of the latest QA datasets were generated using classical tools \cite{tovchigrechko2006gramm, pierce2011accelerating} and, as such, typically contain only a few near-native decoys for each protein target. Consequently, EMA methods trained using these classical datasets may not be complementary to high-accuracy protein complex structure predictors.

\begin{figure}
    \centering
    \includegraphics[width=\textwidth]{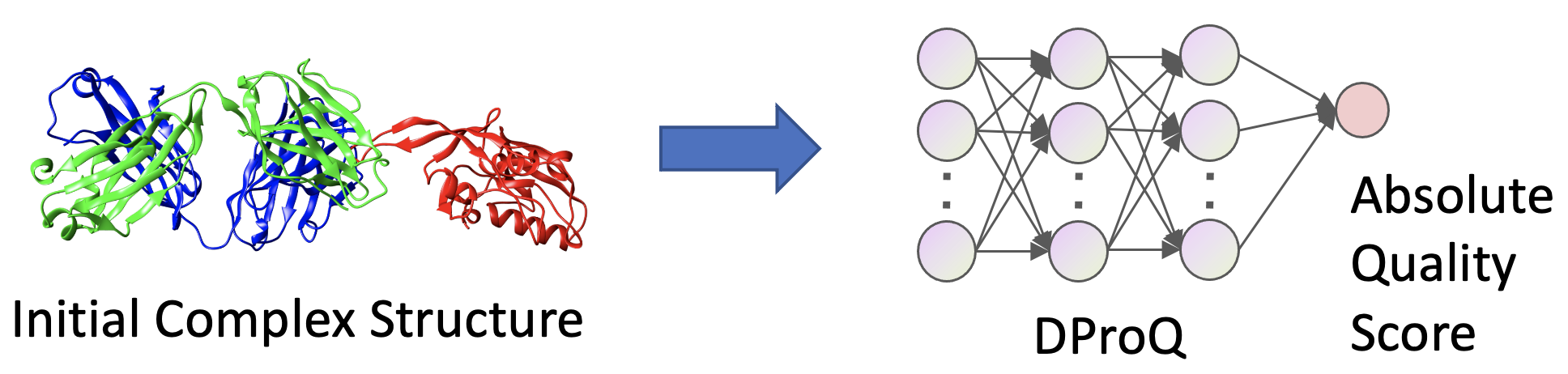}
    \caption{An overview of the protein structure quality assessment problem addressed by \textsc{DProQ}.}
    \label{fig:problem_overview}
\end{figure}

Here, we introduce the Deep Protein Quality (\textsc{DProQ}) for structural ranking of protein complexes - Figure \ref{fig:problem_overview}. In particular, \textsc{DProQ} introduces the Gated Graph Transformer, a novel graph neural network (GNN) that learns to modulate its input information to better guide its structure quality predictions. Being that \textsc{DProQ} requires only a single forward pass to finalize its predictions, we achieve \textbf{prediction time speed-ups} compared to existing QA solutions. Moreover, our model poses structure quality assessment as a multi-task problem, making it the first of its kind in deep learning (DL)-based structure assessment.

\section{Related Work}
We now proceed to describe prior works relevant to DL-based structure quality assessment.\par

\textbf{Biomolecular structure prediction.} Predicting biomolecular structures has been an essential problem for the last several decades. However, very recently, the problem of protein tertiary structure prediction has largely been solved by new DL methods \cite{jumper2021highly, baek2021accurate}. Furthermore, \cite{bryant2022improved} and others have begun making advancements in protein complex structure prediction. Such structure prediction methods have notably accelerated the process of determining 3D protein structures. As such, modern DL-based structure predictors have benefited many related research areas such as drug discovery \cite{macalino2018evolution} and protein design \cite{jendrusch2021alphadesign}.\par

\textbf{Protein representation learning.} Protein structures can be represented in various ways. Previously, proteins have been represented as tableau data in the form of hand-crafted features \cite{chen2020deep}. Along this line, many works \cite{wu2021deepdist, chen2022distema} have represented proteins using pairwise information embeddings such as residue-residue distance maps and contact maps. Recently, describing proteins as graphs has become a popular means of representing proteins, as such representations can learn and leverage proteins' geometric information more naturally. For example, EnQA \cite{chen20223d} used equivariant graph representations to estimate the per-residue quality of protein structures. Moreover, the Equivariant Graph Refiner (EGR) model formulated structural refinement and assessment of protein complexes using semi-supervised equivariant graph neural networks.\par

\textbf{Expert techniques for protein structure quality assessment.} Over the past few decades, many EMA methods have been developed to solve the challenging task of QA \cite{gray2003protein, huang2008iterative, vreven2011integrating, basu2016finding, geng2020iscore}. Among these scoring methods, machine learning-based EMA methods have shown better performance than physics-based \cite{dominguez2003haddock, moal2013scoring} and statistics-based methods \cite{zhou2002distance, pons2011scoring}.\par

\textbf{Machine learning for protein structure quality assessment.} Newly-released machine learning methods have utilized various techniques and features to approach the task of structural QA. For example, ProQDock \cite{basu2016finding} and iScore \cite{geng2020iscore} used protein structure data as the input for a support vector machine classifier. Similarly, EGCN \cite{cao2020energy} assembled graph pairs to represent protein-protein structures and then employed a graph convolutional network (GCN) to learn graph structural information. DOVE \cite{wang2020protein} used a 3D convolutional neural network (CNN) to describe protein-protein interfaces as input features. In a similar spirit, GNN\_DOVE \cite{wang2021protein} and PPDocking \cite{hanquality} trained Graph Attention Networks \cite{velivckovic2017graph} to evaluate protein complex decoys. Moreover, PAUL \cite{eismann2021hierarchical} used a rotation-equivariant neural network to identify accurate models of protein complexes at an atomic level.\par

\textbf{Deep representation learning with Transformers.} Increasingly more works have applied in different domains Transformer-like architectures or multi-head attention (MHA) mechanisms to achieve state-of-the-art (SOTA) results. For example, the Swin-Transformer achieved SOTA performance in various computer vision tasks \cite{hu2019local, liu2021swin, liu2021swinv2}. Likewise, the MSA Transformer \cite{rao2021msa} used tied row and column MHA to address many important tasks in computational biology. Moreover, DeepInteract\cite{morehead2021geometric} introduced the Geometric Transformer to model protein chains as graphs for protein interface contact prediction.\par

\textbf{Contributions.} Our work builds upon prior works by making the following contributions: 

\begin{enumerate}
\item We present the new \textsc{DProQ} pipeline, trained using our newly-developed protein complex datasets in which all structural decoys were generated using AlphaFold2 \cite{jumper2021highly} and AlphaFold-Multimer \cite{evans2021protein}.
\item Using our newly-developed Heterodimer-AF2 (HAF2) and Docking Benchmark 5.5-AF2 (DMB55-AF2) datasets, we demonstrate the superiority of \textsc{DProQ} compared to GNN\_DOVE, the current state-of-the-art for structural QA of protein complexes.
\item We provide the \textit{first} example of applying Transformer representation learning to the task of protein structure quality assessment, by introducing the new Gated Graph Transformer architecture to iteratively update node and edge representations using adaptive feature modulation.
\end{enumerate}

\section{\textsc{DProQ} Pipeline}
\begin{figure}
    \centering
    \includegraphics[width=\textwidth]{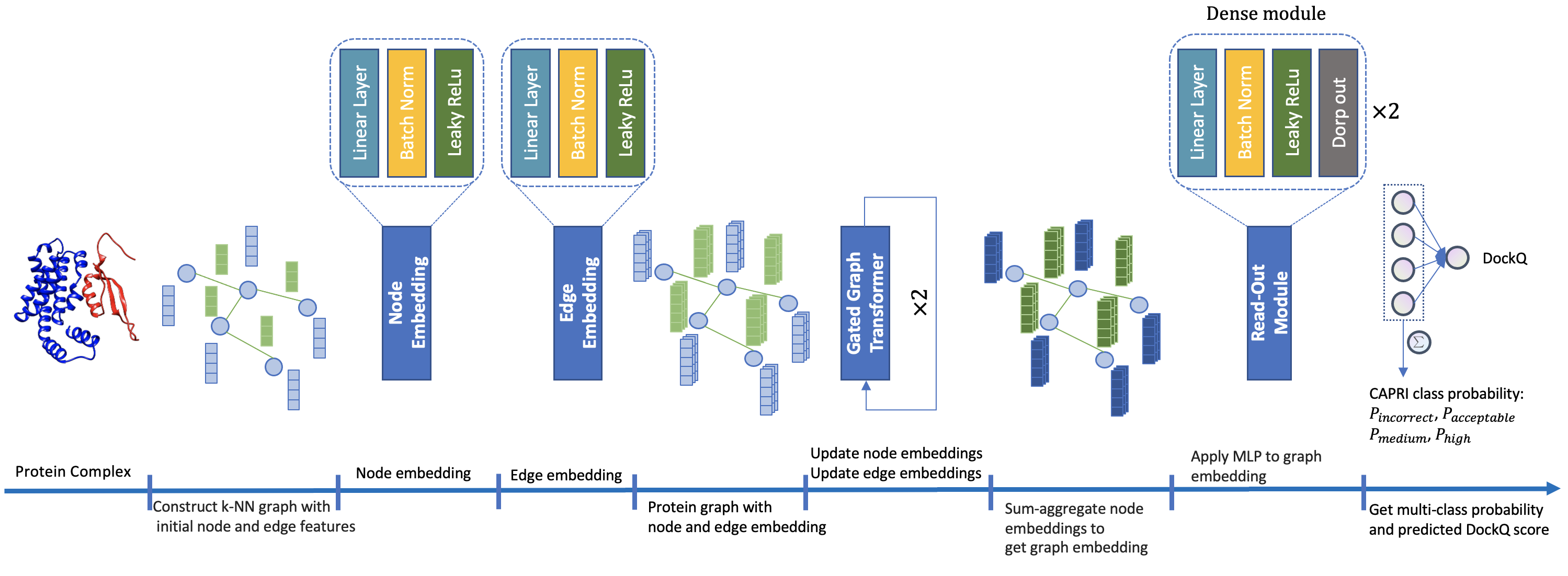}
    \caption{An overview of the \textsc{DProQ} pipeline for quality assessment of protein complexes.}
    \label{fig:pipeline}
\end{figure}

We will now begin to describe \textsc{DProQ}. Shown in Figure \ref{fig:problem_overview} and illustrated from from left to right in Figure \ref{fig:pipeline}, \textsc{DProQ} first receives a 3D protein complex structure as input and represents it as a spatial graph. Notably, all chains in the complex are represented within the same graph topology, where connected pairs of atoms from the same chain are distinguished using a binary edge feature, as described in the following sections. \textsc{DProQ} models protein complexes in this manner to facilitate explicit information flow between chains, which has proven useful for other tasks on macromolecular structures \cite{ganea2021independent, stark2022equibind}. We note that \textsc{DProQ} is not trained using \textit{any} coevolutionary features, making it an end-to-end geometric deep learning method for protein complex structures.
 
\textbf{K-NN graph representation.} \textsc{DProQ} represents each input protein complex structure as a spatial k-nearest neighbors (k-NN) graph $\mathcal{G} = (\mathcal{V}, \mathcal{E})$, where the protein's C$\alpha$ atoms serve as $\mathcal{V}$ (i.e., the nodes of $\mathcal{G}$). After constructing $\mathcal{G}$ by connecting each node to its 10 closest neighbors in $\mathbb{R}^{3}$, we denote its initial node features as $\mathbf{F} \in \mathbb{R}^{d \times N}$ (e.g., residue type) and its initial edge features as $\mathbf{E} \in \mathbb{R}^{d \times M}$ (e.g., C$\alpha$-C$\alpha$ distance), as described more thoroughly in Appendix \ref{sec:appendix_b}.\par

\textbf{Node and edge embeddings.} After receiving a protein complex graph $\mathcal{G}$ as input, \textsc{DProQ} applies initial node and edge embedding modules to each node and edge, respectively. We define such embedding modules as $Embed(\mathbf{F}) = \varphi^{h}(\mathbf{f}_{i}), \forall(i) \in \mathcal{V}$ and $Embed(\mathbf{E}) = \varphi^{e}(\mathbf{f}_{j \rightarrow i}), \forall(i, j) \in \mathcal{E}$, respectively, where each $\varphi$ function is represented as a shallow neural network consisting of a linear layer, batch normalization, and LeakyReLU activation \cite{xu2015empirical}. Such node and edge embeddings are then fed as an updated input graph into the new Gated Graph Transformer.

\subsection{Gated Graph Transformer architecture}
\label{sec:gate_gt}
\begin{figure}
  \centering
  \includegraphics[width=\textwidth]{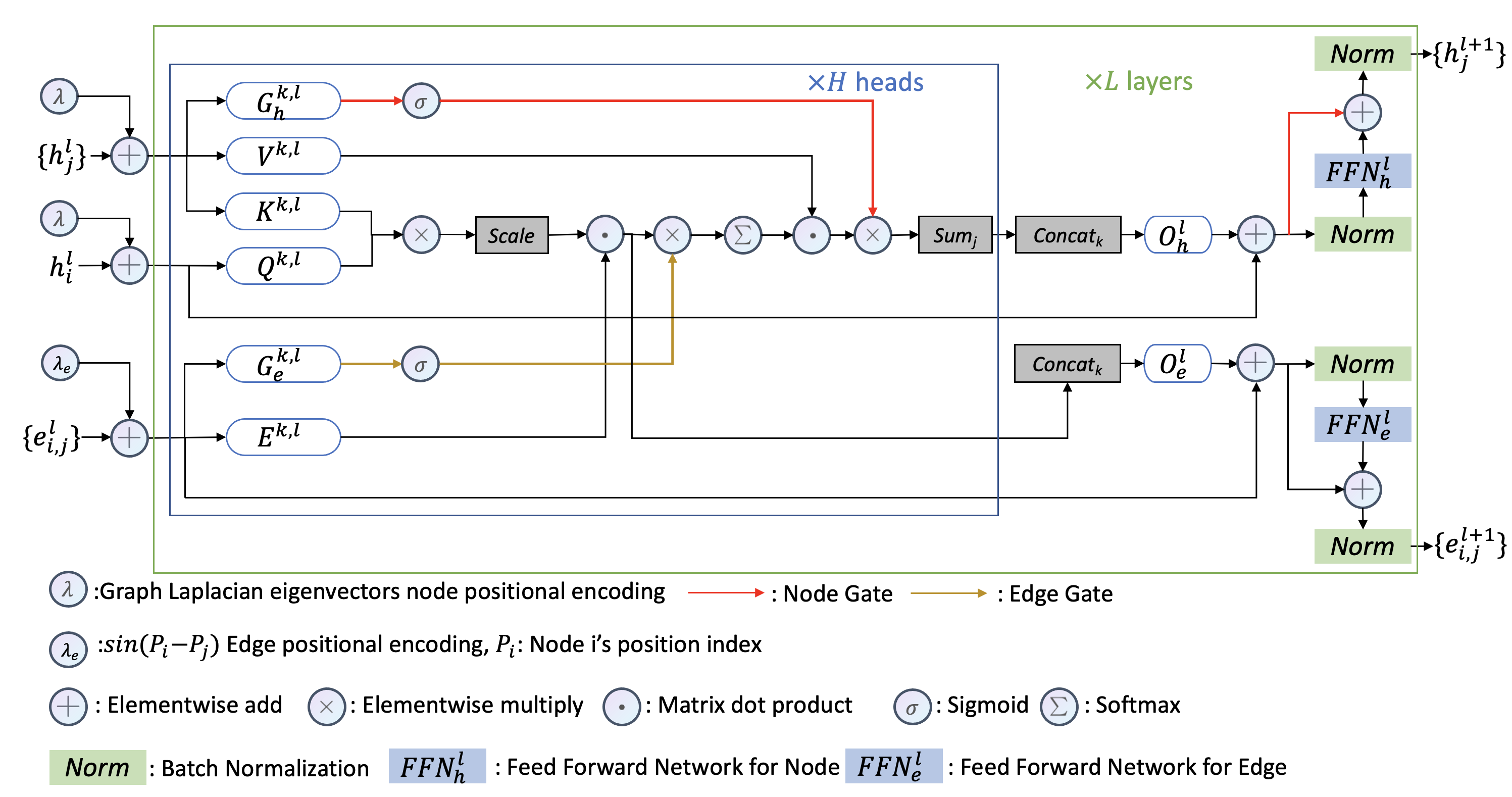}
  \caption{\textsc{GGT} model architecture.}
  \label{fig:gate_gt}
\end{figure}

We designed the Gated Graph Transformer (\textsc{GGT}) as a solution to a specific phenomenon in graph representation learning. Note that, unlike other GNN-based structure scoring methods, \cite{wang2021protein, hanquality} define edges using a fixed distance threshold, which means that each graph node may have a different number of incoming and outgoing edges. In contrast, \textsc{DProQ} constructs and operates on k-NN graphs where all nodes are connected to the same number of neighbors. However, in the context of k-NN graphs, each neighbor's information is, by default, given equal priority during information updates. As such, we may desire to imbue our graph neural network with the ability to automatically decide the priority of different nodes and edges during the graph message passing. Consequently, we present the \textsc{GGT}, a gated neighborhood-modulating Graph Transformer inspired by \cite{velivckovic2017graph, dwivedi2020generalization, morehead2021geometric}. Formally, to update the network's node embeddings $\mathbf{h}_{i}$ and edge embeddings $\mathbf{e}_{ij}$, we define a single layer of the \textsc{GGT} as:

\begin{equation}
\label{eq:w2}
    \widehat{\widehat{\mathbf{w}}}_{ij}^{k, \ell}=\left(\frac{Q^{k, \ell} \mathbf{h}_{i}^{\ell} \cdot K^{k, \ell} \mathbf{h}_{j}^{\ell}}{\sqrt{d_{k}}}\right) \cdot E^{k, \ell} \mathbf{e}_{i j}^{\ell}
\end{equation}

\begin{equation}
    \label{eq:e_update}
    \hat{\mathbf{e}}_{ij}^{\ell+1}=\mathbf{O}_{e}^{\ell} \|_{k=1}^{H}\left(\widehat{\widehat{\mathbf{w}}}_{ij}^{k, \ell}\right)
\end{equation}

\begin{equation}
    \label{eq:w1}
    \widehat{\mathbf{w}}_{ij}^{k, \ell}=\widehat{\widehat{\mathbf{w}}}_{ij}^{k, \ell} \times \operatorname{sigmoid}\left(\mathbf{G}_{e}^{k, \ell} \mathbf{e}_{i j}^{\ell}\right)
\end{equation}

\begin{equation}
    \label{eq:w}
    \mathbf{w}_{ij}^{k, \ell}=\operatorname{softmax}\left(\hat{\mathbf{w}}_{ij}^{k, \ell}\right)
\end{equation}

\begin{equation}
    \label{eq:h_update}
    \hat{\mathbf{h}}_{i}^{\ell+1}=O_{\mathbf{h}}^{\ell} \|_{k=1}^{H}\left(\operatorname{sigmoid}\left(G_{h}^{k, \ell} \mathbf{h}_{j}^{\ell}\right) \times \sum_{j \in \mathcal{N}_{i}} \mathbf{w}_{i j}^{k, \ell} V^{k, \ell} \mathbf{h}_{j}^{\ell}\right)
\end{equation}

In particular, the \textsc{GGT} introduces to the standard Graph Transformer architecture \cite{dwivedi2020generalization} two information gates through which the network can modulate node and edge information flow, as shown in Figure \ref{fig:gate_gt}. Equation \ref{eq:e_update} shows how the output of the edge gate, $\mathbf{G}_{e}$, is used elementwise to gate the network's attention scores $\widehat{\widehat{\mathbf{w}}}_{ij}$ derived from Equation \ref{eq:w2}. Such gating simultaneously allows the network to decide how much to regulate edge information flow as well as how much to weigh the edge of each node neighbor. Similarly, the node gate, $\mathbf{G}_{h}$, shown in Equation \ref{eq:h_update}, allows the GGT to individually decide how to modulate information from node neighbors during node information updates. Lastly, for the GGT's Feed Forward Network, we keep the same structure as described in \cite{dwivedi2020generalization}, providing us with new node embeddings $\hat{\mathbf{h}}_{i}^{L}$ and edge embeddings $\hat{\mathbf{e}}_{ij}^{L}$. For additional background on the GGT's network operations, we refer readers to \cite{dwivedi2020generalization}.

\subsection{Multi-task graph property prediction}
\label{sec:graph_predictions}
To obtain graph-level predictions for each input protein complex graph, we apply a graph sum-pooling operator on $\hat{\mathbf{h}}_{i}^{L}$ such that $\mathbf{p} \in \mathbb{R}^{d \times 1} = \sum_{i = 1}^{N} \mathbf{h}_{i}^{L}$. This graph embedding $\mathbf{p}$ is then fed as input to \textsc{DProQ}'s first read-out module $\varphi_{1}^{r}$, where each read-out module consists of a series of linear layers, batch normalization layers, LeakyReLU activations, and dropout layers \cite{hinton2012improving}, respectively. That is, $\mathbf{\hat{p}} \in \mathbb{R}^{d \times 4} = \varphi_{1}^{r}(\mathbf{p})$ reduces the dimensionality of $\mathbf{p}$ to accommodate \textsc{DProQ}'s output head designed for graph classification. Specifically, for graph classification, we apply a Softmax layer such that $\mathbf{y} \in \mathbb{R}^{d \times 4} = softmax(\mathbf{\hat{p}})$, where $\mathbf{y}$ is the network's predicted DockQ structure quality class probabilities \cite{basu2016dockq} for each input protein complex (e.g., Medium, High). Thereafter, we apply \textsc{DProQ}'s second read-out module $\varphi_{2}^{r}$ to $\mathbf{\hat{p}}$ such that $\mathbf{q} \in \mathbb{R}^{1 \times 1} = \varphi_{2}^{r}(\mathbf{\hat{p}})$. Within this context, the network's scalar graph regression output $\mathbf{q}$ represents the network's predicted DockQ score \cite{basu2016dockq} for a given protein complex input.

\textbf{Structure quality scoring loss.} To train \textsc{DProQ}'s graph regression head, we used the mean squared error loss $\mathcal{L}_{R} = \frac{1}{N} \sum_{i = 1}^{N} \| \mathbf{q}_{i}' - \mathbf{q}_{i}^{*} \|^{2}$. Here, $\mathbf{q}_{i}'$ is the model's predicted DockQ score for example $i$, $\mathbf{q}_{i}^{*}$ is the ground truth DockQ score for example $i$, and $N$ represents the number of examples in a given mini-batch.

\textbf{Structure quality classification loss.} Likewise, to train \textsc{DProQ}'s graph classification head, we used the cross-entropy loss $\mathcal{L}_{C} = \frac{1}{N} \sum_{i = 1}^{N} \left(\mathbf{y}_{i}^{*} - \log(\mathbf{y}_{i}')\right)$. Here, $\mathbf{y}_{i}'$ is the model's predicted DockQ quality class (e.g., Acceptable) for example $i$, and $\mathbf{y}_{i}^{*}$ is the ground truth DockQ quality class (e.g., Incorrect) for example $i$.

\textbf{Overall loss.} We define \textsc{DProQ}'s overall loss as $\mathcal{L} = w_{\mathcal{L}_{C}} \times \mathcal{L}_{C} + w_{\mathcal{L}_{R}} \times \mathcal{L}_{R}$. We note that the weights for each constituent loss (e.g., $w_{\mathcal{L}_{R}}$) were determined either by performing a grid search or instead by using a lowest validation loss criterion for parameter selection. In Appendix \ref{sec:appendix_c}, we describe \textsc{DProQ}'s choice of hyperparameters in greater detail.

\section{Experiments}
\subsection{Data}
Here, we briefly describe our four new datasets for complex structural QA, datasets that we used to train, validate, and test all \textsc{DProQ} models. Appendix \ref{sec:appendix_b} describes in greater detail how we generated these datasets as well as their corresponding labels \footnote{We make this data and associated scripts available at \href{https://github.com/BioinfoMachineLearning/DProQ}{\texttt{https://github.com/BioinfoMachineLearning/DProQ}}}.

\textbf{Dataset Labels.} As shown in Section \ref{sec:graph_predictions}, \textsc{DProQ} performs two learning tasks simultaneously. In the framework of its graph regression task, \textsc{DProQ} treats DockQ scores \cite{basu2016dockq} as its per-graph labels. As introduced earlier, DockQ scores are robust, continuous values in the range of [0, 1]. Such scores measure the quality of a protein complex structure such that a higher DockQ score indicates a higher-quality structure. In the context of its graph classification task, \textsc{DProQ} predicts the probabilities that the structure of an input protein complex falls into the Incorrect, Acceptable, Medium, or High-quality category, where assignments into such quality categories are made according to the structure's true DockQ score.\par

\textbf{Multimer-AF2 Dataset.} To build our first new training and validation dataset, the Multimer-AF2 (MAF2) dataset, we used AlphaFold 2 \cite{jumper2021highly} and AlphaFold-Multimer \cite{evans2021protein} to generate protein complex structures for protein targets derived from the EVCoupling \cite{hopf2019evcouplings} and DeepHomo \cite{yan2021accurate} datasets. In summary, the MAF2 dataset contains a total of $9,251$ decoys, where according to DockQ scores $20.44\%$ of them are of Incorrect quality, $14.34\%$ of them are of Acceptable quality, $30.00\%$ of them are of Medium quality, and the remaining $35.22\%$ of them are of High quality.\par

\textbf{Docking Decoy Set.} Our second new training and validation dataset, the Docking Decoy dataset \cite{kundrotas2018dockground}, contains $58$ protein complex targets. Each target includes approximately $100$ Incorrect decoys and at least one \textit{near-native} decoy. To construct this dataset, we follow the sequence clustering results from GNN\_DOVE \cite{wang2021protein}, select as an optional testing dataset the fourth clustering fold consisting of $14$ targets, and use the remaining $3$ folds for training and validation, where such training and validation splits consist of $34$ targets and $10$ targets, respectively.\par

\textbf{Heterodimer-AF2 Dataset.} We built our first new test dataset, the Heterodimer-AF2 (HAF2) dataset, by collecting the structures of heterodimers from the Protein Data Bank \cite{berman2000protein}. Thereafter, we used a custom, in-house AlphaFold-Multimer \cite{evans2021protein} system to generate different models for each protein heterodimer, subsequently applying 40\% sequence identity filtering to each split for this dataset. Overall, the HAF2 dataset consists of a total of $13$ targets containing $1,849$ decoys, where $33.36\%$ of these decoys are of Incorrect quality, $6.13\%$ of them are of Acceptable quality, $35.62\%$ of them are of Medium quality, and the remaining $24.58\%$ of them are of High quality.\par

\textbf{Docking Benchmark5.5 AF2 Dataset.} To construct our final test dataset, the Docking Benchmark 5.5-AF2 (DBM55-AF2) dataset, we applied AlphaFold-Multimer \cite{evans2021protein} to predict the structures of Docking Benchmark 5.5 targets \cite{vreven2015updates}. This dataset contains a total of $15$ protein targets comprised of $449$ decoys. $50.78\%$ of these decoys are of Incorrect quality, $16.70\%$ of them are of Acceptable quality, $30.73\%$ of them are of Medium quality, and the remaining $1.78\%$ of them are of High quality.\par
 
\textbf{Cross-Validation Datasets and Overlap Reduction.} All \textsc{DProQ} models were trained on $8,733$ decoys and validated on $3,407$ decoys derived as a combination of our MAF2 and Docking Decoy datasets. After training, each model was blindly tested on our HAF2 and DBM55-AF2 test datasets. To prevent data leakage between our training and test datasets, we used MMseqs2 \cite{ulirdita2021fast} to perform $30\%$ sequence identity filtering w.r.t. our test dataset and training dataset splits. After filtering, we further remove any targets that do not contain any Acceptable or higher-quality decoys according to DockQ's score classifications \cite{basu2016dockq}.\par

\subsection{Evaluation Setup}
\textbf{Baselines.} We compare our method with the state-of-the-art method GNN\_DOVE, an atom-level graph attention-based method for protein complex structure evaluation. Uniquely, it extracts protein interface areas to build its input graphs. Further, the chemical properties of its atoms as well as inter-atom distances are treated as initial node and edge features, respectively. We note that GNN\_DOVE is the sole baseline we include in this study, as for all other DL-based methods for structural QA, we were unable to locate reproducible training and inference code for such methods. Furthermore, since GNN\_DOVE has previously been evaluated against several other classical machine learning methods for structural QA and has demonstrated strong performance against such methods, we argue that comparing new DL methods' performance to GNN\_DOVE can serve as a strong indicator of the DL state-of-the-art for the field.

\textbf{\textsc{DProQ} Models.} Besides including results for the standard \textsc{DProQ} model as well as for GNN\_DOVE, we also report results on the HAF2 and DBM55-AF2 datasets for a selection of \textsc{DProQ} variants curated in this study. The \textsc{DProQ} variants we chose to investigate in this work include \textsc{DProQ\_GT} which employs the original Graph Transformer architecture \cite{dwivedi2020generalization}; \textsc{DProQ\_GTE} which employs the GGT with only its edge gate enabled; and \textsc{DProQ\_GTN} which employs the GGT with only its node gate enabled.\par

\textbf{Evaluation metrics.} We evaluate structural QA performance according to two main criteria. Our first criterion is general ranking ability which measures how many qualified decoys are found within a model's predicted Top-$N$ structure ranking. Within this framework, a model's hit rate is defined as the fraction of protein complex targets for which the model, within each of its Top-$N$ ranks, ranked at least one Acceptable or higher-quality decoy based on the CAPRI scoring criteria \cite{lensink2014score_set}. In this work, we report results based on models' Top-10 hit rates. A hit rate is represented by three numbers separated by the character /. These three numbers, in order, represent how many decoys with Acceptable or higher-quality, Medium or higher-quality, and High quality were among the Top-N ranked decoys. Our second criterion with which to score models is their Top-$1$ selected model ranking ability. Within this context, we calculate the ranking loss for each method. Here, per-target ranking loss is defined as the difference between the DockQ score of a target's native structure and the DockQ score of the top decoy predicted by each ranking method. As such, a lower ranking loss indicates a stronger model for ranking pools of structural decoys for downstream tasks.

\textbf{Implementation Details.} We train all our models using the AdamW optimizer \cite{loshchilov2017decoupled} and perform early stopping with a patience of $15$ epochs. All remaining hyperparameters and graph features we used are described further in Appendix \ref{sec:appendix_c}. Source code to reproduce our results and perform fast structure quality assessment using our model weights can be found at \href{https://github.com/BioinfoMachineLearning/DProQ}{\texttt{https://github.com/BioinfoMachineLearning/DProQ}}.

\subsection{Results}
\textbf{Blind Structure Quality Assessment on the HAF2 Dataset.} Table \ref{tab:haf2_hits} reports the hit rate performances of \textsc{DProQ}, \textsc{DProQ} variants, and GNN\_DOVE, respectively, on the HAF2 Dataset. On $13$ HAF2 targets, \textsc{DProQ} hits (i.e., identifies at least one suitable decoy structure for) $10$ targets on the Acceptable quality level, $9$ targets on the Medium quality level, and $4$ targets on High quality level. For targets with High quality structures, \textsc{DProQ} and its variants successfully hit all of them. On the HAF2 dataset, GNN\_DOVE gets a $8/7/3$ hit rate on Acceptable, Medium, and High quality targets respectively. However, \textsc{DProQ} outperforms GNN\_DOVE on all targets except \textsc{7NKZ}. On this target, GNN\_DOVE achieves a better hit rate on High quality decoy structures. Overall, \textsc{DProQ} shows a better performance than \textsc{DProQ\_GT}, \textsc{DProQ\_GTE}, and \textsc{DProQ\_GTN}. Additionally, for target \textsc{7MRW}, only \textsc{DProQ} successfully hits the $5$ Acceptable quality targets and $4$ Medium quality targets, whereas all other methods failed to hit any qualified decoy structures for this protein target.

Table \ref{tab:haf2_ranking} reports methods' ranking loss performance on each HAF2 target. \textsc{DProQ} and its $3$ variants outperform GNN\_DOVE which displays the highest average ranking loss. In particular, \textsc{DProQ} achieves the lowest average ranking loss of $0.195$ compared to all other methods. Here, \textsc{DProQ}'s loss is $43\%$ lower than GNN\_DOVE's $0.343$ average ranking loss. Moreover, \textsc{DProQ} achieves the lowest loss on $5$ targets, namely \textsc{7AWV}, \textsc{7OEL}, \textsc{7NRW}, \textsc{7NKZ} and \textsc{7O27}. Notably, \textsc{DProQ\_GT}, \textsc{DProQ\_GTE}, and \textsc{DProQ\_GTN}'s ranking losses here are $29\%$, $34\%$, and $30\%$ lower, respectively, than GNN\_DOVE's average ranking loss.

\begin{table}
\caption{Hit rate performance on the HAF2 dataset. The \textsc{Best} column represents each target's best-possible Top-$10$ result. The \textsc{Summary} row lists the results when all targets are taken into consideration.}
\label{tab:haf2_hits}
\centering
\begin{tabular}{lllllll}
\toprule
ID      & \textsc{DProQ}    & \textsc{DProQ}\_GT & \textsc{DProQ}\_GTE & \textsc{DProQ}\_GTN  & GNN\_DOVE & \textsc{BEST} \\ 
\midrule
7AOH    & 10/10/10 & 10/10/10  & 10/10/10   & 10/10/10        & 9/9/0     & 10/10/10   \\
7D7F    & 0/0/0    & 0/0/0     & 0/0/0      & 0/0/0           & 0/0/0     & 5/0/0      \\ 
7AMV    & 10/10/10 & 10/10/10  & 10/10/10   & 10/10/10        & 10/10/6   & 10/10/10   \\ 
7OEL    & 10/10/0  & 10/9/0    & 10/10/0    & 10/10/0         & 10/10/0   & 10/10/0    \\ 
7O28    & 10/10/0  & 10/10/0   & 10/10/0    & 10/10/0         & 10/10/0   & 10/10/0    \\ 
7ALA    & 0/0/0    & 0/0/0     & 0/0/0      & 0/0/0           & 0/0/0     & 1/0/0      \\ 
7MRW    & 5/4/0    & 0/0/0     & 0/0/0      & 0/0/0           & 0/0/0     & 10/10/0    \\ 
7OZN    & 0/0/0    & 0/0/0     & 0/0/0      & 0/0/0           & 0/0/0     & 10/2/0     \\ 
7D3Y    & 2/0/0    & 5/0/0     & 6/0/0      & 8/0/0           & 0/0/0     & 10/0/0     \\ 
7NKZ    & 10/10/2  & 10/10/1   & 10/10/1    & 10/010/4        & 10/9/9    & 10/10/10   \\ 
7LXT    & 1/1/0    & 0/0/0     & 0/0/0      & 0/0/0           & 1/0/0     & 10/10/0    \\ 
7KBR    & 10/10/10 & 10/10/10  & 10/10/10   & 10/10/10        & 10/10/9   & 10/10/10   \\ 
7O27    & 10/10/0  & 10/10/0   & 10/10/0    & 10/10/0         & 10/4/0    & 10/10/0    \\ 
\midrule
\textsc{Summary} & \textbf{10/9/4} & 8/7/4 & 8/7/4   & 8/7/4           & 8/7/3     & 13/10/4    \\ 
\bottomrule
\end{tabular}
\end{table}

\begin{table}
\caption{Ranking loss performance on the HAF2 dataset. The \textsc{BEST} row represents the mean and standard deviation of the ranking losses for all targets.}
\label{tab:haf2_ranking}
\centering
\begin{tabular}{llllll}
\toprule
Target  & \textsc{DProQ}         & DProQ\_GT     & \textsc{DProQ}\_GTE             & \textsc{DProQ}\_GTN    & GNN\_DOVE     \\
\midrule
7AOH    & 0.066         & 0.026         & 0.026                  & 0.058         & 0.928         \\
7D7F    & 0.471         & 0.471         & 0.47                   & 0.471         & 0.003         \\
7AMV    & 0.01          & 0.021         & 0.017                  & 0.019         & 0.342         \\
7OEL    & 0.062         & 0.063         & 0.135                  & 0.135         & 0.21          \\
7O28    & 0.029         & 0.021         & 0.027                  & 0.034         & 0.244         \\
7ALA    & 0.232         & 0.226         & 0.226                  & 0.226         & 0.226         \\
7MRW    & 0.085         & 0.603         & 0.555                  & 0.555         & 0.598         \\
7OZN    & 0.409         & 0.409         & 0.49                   & 0.281         & 0.457         \\
7D3Y    & 0.326         & 0.33          & 0.012                  & 0.326         & 0.295         \\
7NKZ    & 0.164         & 0.175         & 0.175                  & 0.164         & 0.459         \\
7LXT    & 0.586         & 0.586         & 0.586                  & 0.586         & 0.295         \\
7KBR    & 0.068         & 0.152         & 0.152                  & 0.17          & 0.068         \\
7O27    & 0.03          & 0.079         & 0.079                  & 0.079         & 0.334         \\
\midrule
\textsc{BEST} & \textbf{0.195 ± 0.185} & 0.243 ± 0.206 & 0.227 ± 0.21 & 0.239 ± 0.187 & 0.343 ± 0.228   \\
\bottomrule
\end{tabular}
\end{table}

\textbf{Blind Structure Quality Assessment on the DBM55-AF2 Dataset.} Table \ref{tab:dbm_hits} summarizes all methods' hit rate results on the DBM55-AF2 dataset, which contains $15$ targets. Here, we see that \textsc{DProQ} achieves the best hit rate of $12/10/3$ compared to all other methods. Additionally, \textsc{DProQ} successfully selects all $10$ targets with Medium quality decoys and all $3$ targets with High quality decoys, moreover hitting all three accuracy levels' decoys on $6$ targets which is the best result achieved compared to other methods. On this dataset, GNN\_DOVE achieves a $10/4/1$ hit rate, worse than all other methods. Nonetheless, GNN\_DOVE successfully hits two targets, \textsc{5WK3} and \textsc{5KOV}.

Table \ref{tab:dbm_ranking} presents the ranking loss for all methods on DBM55-AF2 dataset. Here, \textsc{DProQ} achieves the best ranking loss of $0.049$ which is $87\%$ lower than GNN\_DOVE's ranking loss of $0.379$. Furthermore, for $4$ targets, \textsc{DProQ} correctly selects the Top-$1$ model and achieves $0$ ranking loss. Similarly to their results on the HAF2 dataset, \textsc{DProQ\_GT}, \textsc{DProQ\_GTE}, and \textsc{DProQ\_GTN}'s losses are $71\%$, $84\%$, and $74\%$ lower, respectively, than GNN\_DOVE's ranking loss.

\begin{table}
\label{tab:dbm_hits}
\centering
\caption{Hit rate performance on DBM55-AF2 dataset. The \textsc{BEST} column represents each target's best-possible Top-$10$ result. The \textsc{Summary} row lists the results when all targets are taken into consideration.}
\begin{tabular}{lllllll}
\toprule
Target   & \textsc{DProQ} & \textsc{DProQ}\_GT & \textsc{DProQ}\_GTE & \textsc{DProQ}\_GTN & GNN\_DOVE & \textsc{BEST}   \\ 
\midrule
6AL0     & 9/2/0          & 10/0/0             & 10/0/0              & 10/2/0              & 6/0/0     & 10/2/0  \\
3SE8     & 8/8/0          & 9/9/0              & 8/8/0               & 8/8/0               & 3/0/0     & 10/10/0 \\
5GRJ     & 10/10/0        & 9/9/0              & 10/10/0             & 9/9/0               & 3/2/0     & 10/10/0 \\
6A77     & 7/7/0          & 7/7/0              & 8/8/0               & 8/8/0               & 0/0/0     & 8/8/0   \\
4M5Z     & 10/10/1        & 10/10/0            & 10/10/0             & 10/10/0             & 10/10/0   & 10/10/1 \\
4ETQ     & 1/1/0          & 1/1/0              & 1/1/0               & 1/1/0               & 0/0/0     & 1/1/0   \\
5CBA     & 10/10/1        & 10/10/0            & 10/10/0             & 10/10/1             & 10/10/3   & 10/10/6 \\
5WK3     & 0/0/0          & 0/0/0              & 0/0/0               & 0/0/0               & 1/0/0     & 3/0/0   \\
5Y9J     & 4/0/0          & 6/0/0              & 5/0/0               & 4/0/0               & 0/0/0     & 8/0/0   \\
6BOS     & 10/10/0        & 10/10/0            & 10/10/0             & 10/10/0             & 10/10/0   & 10/10/0 \\
5HGG     & 8/0/0          & 8/0/0              & 8/0/0               & 8/0/0               & 8/0/0     & 10/0/0  \\
6A0Z     & 0/0/0          & 0/0/0              & 0/0/0               & 0/0/0               & 2/0/0     & 3/0/0   \\
3U7Y     & 2/2/1          & 2/2/1              & 2/2/1               & 2/1/0               & 2/2/1     & 2/2/1   \\
3WD5     & 10/8/0         & 9/8/0              & 9/8/0               & 9/8/0               & 0/0/0     & 10/10/0 \\
5KOV     & 0/0/0          & 0/0/0              & 0/0/0               & 0/0/0               & 1/0/0     & 2/0/0   \\
\midrule
\textsc{Summary}  & \textbf{12/10/3} & 12/9/1           & 12/9/1              & 12/10/1             & 10/4/1    & 15/10/3 \\ 
\bottomrule
\end{tabular}
\end{table}

\begin{table}
\label{tab:dbm_ranking}
\centering
\caption{Ranking loss performance on the DBM55-AF2 dataset. The \textsc{BEST} row represents the mean and standard deviation of the ranking losses for all targets.}
\begin{tabular}{llllll}
\toprule
Target  & \textsc{DProQ}         & \textsc{DProQ}\_GT     & \textsc{DProQ}\_GTE             & \textsc{DProQ}\_GTN    & GNN\_DOVE     \\
\midrule
6AL0    & 0.0           & 0.156         & 0.156                  & 0.0           & 0.424         \\
3SE8    & 0.079         & 0.041         & 0.041                  & 0.079         & 0.735         \\
5GRJ    & 0.024         & 0.012         & 0.095                  & 0.012         & 0.776         \\
6A77    & 0.037         & 0.062         & 0.0                    & 0.037         & 0.591         \\
4M5Z    & 0.015         & 0.026         & 0.026                  & 0.015         & 0.221         \\
4ETQ    & 0.0           & 0.76          & 0.0                    & 0.748         & 0.759         \\
5CBA    & 0.052         & 0.038         & 0.052                  & 0.058         & 0.019         \\
5WK3    & 0.114         & 0.114         & 0.114                  & 0.186         & 0.087         \\
5Y9J    & 0.0           & 0.0           & 0.0                    & 0.0           & 0.382         \\
6BOS    & 0.081         & 0.081         & 0.0                    & 0.0           & 0.081         \\
5HGG    & 0.051         & 0.051         & 0.121                  & 0.051         & 0.121         \\
6A0Z    & 0.207         & 0.207         & 0.207                  & 0.207         & 0.062         \\
3U7Y    & 0.0           & 0.021         & 0.0                    & 0.0           & 0.756         \\
3WD5    & 0.011         & 0.011         & 0.011                  & 0.0           & 0.672         \\
5KOV    & 0.065         & 0.08          & 0.085                  & 0.087         & 0.0           \\
\midrule
\textsc{BEST} & \textbf{0.049 ± 0.054} & 0.111 ± 0.182 & 0.061 ± 0.064 & 0.099 ± 0.185 & 0.379 ± 0.298    \\
\bottomrule
\end{tabular}
\end{table}

\textbf{Node and Edge Gates.} In Tables \ref{tab:haf2_hits}, \ref{tab:haf2_ranking}, \ref{tab:dbm_hits}, and \ref{tab:dbm_ranking}, we evaluated \textsc{DProQ\_GT}, \textsc{DProQ\_GTE}, and \textsc{DProQ\_GTN}'s performance on our two test datasets to discover the node and edge gates' effects on performance. On the HAF2 dataset, our results indicate that exclusively adding either a node or edge gate can help improve a model's Top-$1$ structure ranking ability.

Our gate-ablation results yielded another interesting phenomenon when we observed that \textsc{DProQ\_GTE} consistently achieved the best ranking loss on both of our test datasets, compared to \textsc{DProQ\_GTN}. Notably, on the DBM55-AF2 dataset, \textsc{DProQ\_GTE}'s loss is much lower than that of other methods, except for \textsc{DProQ}. This suggests \textsc{DProQ\_GTE} is more sensitive to Top-$1$ structure selection compared to \textsc{DProQ\_GTN}. We hypothesize this trend may be because, within \textsc{DProQ}, k-NN graphs' edges may not always possess a biological interpretation during graph message passing. In such cases, models' edge gates may allow the network to inhibit the weight of biologically-implausible edges, which could remove noise from the neural network's training procedure. We plan to investigate extensions of the GGT to explore these ideas further.

\textbf{Repeated Experiments.} In addition, we trained four more \textsc{DProQ} models within the original \textsc{DProQ} training environment but with different random seeds. We see in Appendix \ref{sec:appendix_a} that \textsc{DProQ} demonstrates consistently-better results compared to GNN\_DOVE. We observe \textsc{DProQ} providing reliable, state-of-the-art performance in terms of both hit rate and ranking loss on both test datasets.\par

\textbf{Limitations.} One limitation of the \textsc{GGT} is that its architecture does not directly operate using protein spatial information. Models such as the SE(3)-Transformer \cite{fuchs2020se} may be one way to explore this spatial information gap. However, due to such models' high computational costs, we are currently unable to utilize such techniques within the context of protein complex structure modeling. Moreover, in our studies, \textsc{DProQ} failed on some test targets that contain few Acceptable or higher-quality decoys within the Top-$10$ range. To overcome this problem, we experimented with multi-tasking within \textsc{DProQ}, but we did not find such multi-tasking to help our models generalize better. We hypothesize that new, diverse datasets may be helpful in this regard. We defer such ideas to future work.

\section{Conclusion}
\label{sec:conclusion}
In this work, we presented \textsc{DProQ} which introduces the Gated Graph Transformer for protein complex structure assessment. Our rigorous experiments demonstrate that \textsc{DProQ} achieves state-of-the-art performance and efficiency compared to all other DL-based structure assessment methods. In this work, we also introduced three new protein complex structure datasets created using AlphaFold 2 and AlphaFold-Multimer, datasets that we have made publicly available for modeling and benchmarking DL methods. Used with care and with feedback from domain experts, we believe \textsc{DProQ} and its datasets may help accelerate the adoption of DL methods in responsible drug discovery.

\section*{Acknowledgments}
The project is partially supported by two NSF grants (DBI 1759934 and IIS 1763246), one NIH grant (GM093123), three DOE grants (DE-SC0020400, DE-AR0001213, and DE-SC0021303), and the computing allocation on the Summit compute cluster provided by Oak Ridge Leadership Computing Facility (Contract No. DE-AC05-00OR22725).

{
\small
\printbibliography

@article{scott2016small,
  title={Small molecules, big targets: drug discovery faces the protein--protein interaction challenge},
  author={Scott, Duncan E and Bayly, Andrew R and Abell, Chris and Skidmore, John},
  journal={Nature Reviews Drug Discovery},
  volume={15},
  number={8},
  pages={533--550},
  year={2016},
  publisher={Nature Publishing Group}
}

@article{athanasios2017protein,
  title={Protein-protein interaction (PPI) network: recent advances in drug discovery},
  author={Athanasios, Alexiou and Charalampos, Vairaktarakis and Vasileios, Tsiamis and others},
  journal={Current drug metabolism},
  volume={18},
  number={1},
  pages={5--10},
  year={2017},
  publisher={Bentham Science Publishers}
}

@article{macalino2018evolution,
  title={Evolution of in silico strategies for protein-protein interaction drug discovery},
  author={Macalino, Stephani Joy Y and Basith, Shaherin and Clavio, Nina Abigail B and Chang, Hyerim and Kang, Soosung and Choi, Sun},
  journal={Molecules},
  volume={23},
  number={8},
  pages={1963},
  year={2018},
  publisher={Multidisciplinary Digital Publishing Institute}
}

@article{kortemme2004computational,
  title={Computational design of protein--protein interactions},
  author={Kortemme, Tanja and Baker, David},
  journal={Current opinion in chemical biology},
  volume={8},
  number={1},
  pages={91--97},
  year={2004},
  publisher={Elsevier}
}

@article{baker2006prediction,
  title={Prediction and design of macromolecular structures and interactions},
  author={Baker, David},
  journal={Philosophical Transactions of the Royal Society B: Biological Sciences},
  volume={361},
  number={1467},
  pages={459--463},
  year={2006},
  publisher={The Royal Society London}
}

@article{lippow2007progress,
  title={Progress in computational protein design},
  author={Lippow, Shaun M and Tidor, Bruce},
  journal={Current opinion in biotechnology},
  volume={18},
  number={4},
  pages={305--311},
  year={2007},
  publisher={Elsevier}
}

@article{evans2021protein,
  title={Protein complex prediction with AlphaFold-Multimer},
  author={Evans, Richard and O'Neill, Michael and Pritzel, Alexander and Antropova, Natasha and Senior, Andrew W and Green, Timothy and {\v{Z}}{\'\i}dek, Augustin and Bates, Russell and Blackwell, Sam and Yim, Jason and others},
  journal={BioRxiv},
  year={2021},
  publisher={Cold Spring Harbor Laboratory}
}

@article{gray2003protein,
  title={Protein--protein docking with simultaneous optimization of rigid-body displacement and side-chain conformations},
  author={Gray, Jeffrey J and Moughon, Stewart and Wang, Chu and Schueler-Furman, Ora and Kuhlman, Brian and Rohl, Carol A and Baker, David},
  journal={Journal of molecular biology},
  volume={331},
  number={1},
  pages={281--299},
  year={2003},
  publisher={Elsevier}
}

@article{vreven2011integrating,
  title={Integrating atom-based and residue-based scoring functions for protein--protein docking},
  author={Vreven, Thom and Hwang, Howook and Weng, Zhiping},
  journal={Protein Science},
  volume={20},
  number={9},
  pages={1576--1586},
  year={2011},
  publisher={Wiley Online Library}
}

@article{huang2008iterative,
  title={An iterative knowledge-based scoring function for protein--protein recognition},
  author={Huang, Sheng-You and Zou, Xiaoqin},
  journal={Proteins: Structure, Function, and Bioinformatics},
  volume={72},
  number={2},
  pages={557--579},
  year={2008},
  publisher={Wiley Online Library}
}

@article{basu2016finding,
  title={Finding correct protein--protein docking models using ProQDock},
  author={Basu, Sankar and Wallner, Bj{\"o}rn},
  journal={Bioinformatics},
  volume={32},
  number={12},
  pages={i262--i270},
  year={2016},
  publisher={Oxford University Press}
}

@article{cao2020energy,
  title={Energy-based graph convolutional networks for scoring protein docking models},
  author={Cao, Yue and Shen, Yang},
  journal={Proteins: Structure, Function, and Bioinformatics},
  volume={88},
  number={8},
  pages={1091--1099},
  year={2020},
  publisher={Wiley Online Library}
}

@article{geng2020iscore,
  title={iScore: a novel graph kernel-based function for scoring protein--protein docking models},
  author={Geng, Cunliang and Jung, Yong and Renaud, Nicolas and Honavar, Vasant and Bonvin, Alexandre MJJ and Xue, Li C},
  journal={Bioinformatics},
  volume={36},
  number={1},
  pages={112--121},
  year={2020},
  publisher={Oxford University Press}
}

@article{wang2020protein,
  title={Protein docking model evaluation by 3D deep convolutional neural networks},
  author={Wang, Xiao and Terashi, Genki and Christoffer, Charles W and Zhu, Mengmeng and Kihara, Daisuke},
  journal={Bioinformatics},
  volume={36},
  number={7},
  pages={2113--2118},
  year={2020},
  publisher={Oxford Academic}
}

@article{dominguez2003haddock,
  title={HADDOCK: a protein- protein docking approach based on biochemical or biophysical information},
  author={Dominguez, Cyril and Boelens, Rolf and Bonvin, Alexandre MJJ},
  journal={Journal of the American Chemical Society},
  volume={125},
  number={7},
  pages={1731--1737},
  year={2003},
  publisher={ACS Publications}
}

@article{moal2013scoring,
  title={The scoring of poses in protein-protein docking: current capabilities and future directions},
  author={Moal, Iain H and Torchala, Mieczyslaw and Bates, Paul A and Fern{\'a}ndez-Recio, Juan},
  journal={BMC bioinformatics},
  volume={14},
  number={1},
  pages={1--15},
  year={2013},
  publisher={BioMed Central}
}

@article{zhou2002distance,
  title={Distance-scaled, finite ideal-gas reference state improves structure-derived potentials of mean force for structure selection and stability prediction},
  author={Zhou, Hongyi and Zhou, Yaoqi},
  journal={Protein science},
  volume={11},
  number={11},
  pages={2714--2726},
  year={2002},
  publisher={Wiley Online Library}
}

@article{pons2011scoring,
  title={Scoring by intermolecular pairwise propensities of exposed residues (SIPPER): a new efficient potential for protein- protein docking},
  author={Pons, Carles and Talavera, David and De La Cruz, Xavier and Orozco, Modesto and Fernandez-Recio, Juan},
  journal={Journal of chemical information and modeling},
  volume={51},
  number={2},
  pages={370--377},
  year={2011},
  publisher={ACS Publications}
}

@article{wang2021protein,
  title={Protein docking model evaluation by graph neural networks},
  author={Wang, Xiao and Flannery, Sean T and Kihara, Daisuke},
  journal={Frontiers in Molecular Biosciences},
  volume={8},
  pages={402},
  year={2021},
  publisher={Frontiers}
}

@article{hanquality,
  title={Quality Assessment of Protein Docking Models Based on Graph Neural Network},
  author={Han, Ye and Chen, Yongbing and He, Fei and Qin, Wenyuan and Yu, Helong and Xu, Dong},
  journal={Frontiers in Bioinformatics},
  pages={30},
  publisher={Frontiers}
}

@article{lensink2014score_set,
  title={Score\_set: a CAPRI benchmark for scoring protein complexes},
  author={Lensink, Marc F and Wodak, Shoshana J},
  journal={Proteins: Structure, Function, and Bioinformatics},
  volume={82},
  number={11},
  pages={3163--3169},
  year={2014},
  publisher={Wiley Online Library}
}

@software{Falcon_PyTorch_Lightning_2019,
author = {Falcon, William and {The PyTorch Lightning team}},
doi = {10.5281/zenodo.3828935},
license = {Apache-2.0},
month = {3},
title = {{PyTorch Lightning}},
url = {https://github.com/PyTorchLightning/pytorch-lightning},
version = {1.4},
year = {2019}
}

@article{basu2016dockq,
  title={DockQ: a quality measure for protein-protein docking models},
  author={Basu, Sankar and Wallner, Bj{\"o}rn},
  journal={PloS one},
  volume={11},
  number={8},
  pages={e0161879},
  year={2016},
  publisher={Public Library of Science San Francisco, CA USA}
}

@article{wang2019dgl,
    title={Deep Graph Library: A Graph-Centric, Highly-Performant Package for Graph Neural Networks},
    author={Minjie Wang and Da Zheng and Zihao Ye and Quan Gan and Mufei Li and Xiang Song and Jinjing Zhou and Chao Ma and Lingfan Yu and Yu Gai and Tianjun Xiao and Tong He and George Karypis and Jinyang Li and Zheng Zhang},
    year={2019},
    journal={arXiv preprint arXiv:1909.01315}
}

@incollection{NEURIPS2019_9015,
title = {PyTorch: An Imperative Style, High-Performance Deep Learning Library},
author = {Paszke, Adam and Gross, Sam and Massa, Francisco and Lerer, Adam and Bradbury, James and Chanan, Gregory and Killeen, Trevor and Lin, Zeming and Gimelshein, Natalia and Antiga, Luca and Desmaison, Alban and Kopf, Andreas and Yang, Edward and DeVito, Zachary and Raison, Martin and Tejani, Alykhan and Chilamkurthy, Sasank and Steiner, Benoit and Fang, Lu and Bai, Junjie and Chintala, Soumith},
booktitle = {Advances in Neural Information Processing Systems 32},
pages = {8024--8035},
year = {2019},
publisher = {Curran Associates, Inc.},
url = {http://papers.neurips.cc/paper/9015-pytorch-an-imperative-style-high-performance-deep-learning-library.pdf}
}

@article{joosten2010series,
  title={A series of PDB related databases for everyday needs},
  author={Joosten, Robbie P and Te Beek, Tim AH and Krieger, Elmar and Hekkelman, Maarten L and Hooft, Rob WW and Schneider, Reinhard and Sander, Chris and Vriend, Gert},
  journal={Nucleic acids research},
  volume={39},
  number={suppl\_1},
  pages={D411--D419},
  year={2010},
  publisher={Oxford University Press}
}

@article{kabsch1983dictionary,
  title={Dictionary of protein secondary structure: pattern recognition of hydrogen-bonded and geometrical features},
  author={Kabsch, Wolfgang and Sander, Christian},
  journal={Biopolymers: Original Research on Biomolecules},
  volume={22},
  number={12},
  pages={2577--2637},
  year={1983},
  publisher={Wiley Online Library}
}

@article{cock2009biopython,
  title={Biopython: freely available Python tools for computational molecular biology and bioinformatics},
  author={Cock, Peter JA and Antao, Tiago and Chang, Jeffrey T and Chapman, Brad A and Cox, Cymon J and Dalke, Andrew and Friedberg, Iddo and Hamelryck, Thomas and Kauff, Frank and Wilczynski, Bartek and others},
  journal={Bioinformatics},
  volume={25},
  number={11},
  pages={1422--1423},
  year={2009},
  publisher={Oxford University Press}
}

@article{dwivedi2020generalization,
  title={A generalization of transformer networks to graphs},
  author={Dwivedi, Vijay Prakash and Bresson, Xavier},
  journal={arXiv preprint arXiv:2012.09699},
  year={2020}
}

@article{hopf2019evcouplings,
  title={The EVcouplings Python framework for coevolutionary sequence analysis},
  author={Hopf, Thomas A and Green, Anna G and Schubert, Benjamin and Mersmann, Sophia and Sch{\"a}rfe, Charlotta PI and Ingraham, John B and Toth-Petroczy, Agnes and Brock, Kelly and Riesselman, Adam J and Palmedo, Perry and others},
  journal={Bioinformatics},
  volume={35},
  number={9},
  pages={1582--1584},
  year={2019},
  publisher={Oxford University Press}
}

@article{yan2021accurate,
  title={Accurate prediction of inter-protein residue--residue contacts for homo-oligomeric protein complexes},
  author={Yan, Yumeng and Huang, Sheng-You},
  journal={Briefings in bioinformatics},
  volume={22},
  number={5},
  pages={bbab038},
  year={2021},
  publisher={Oxford University Press}
}

@article{morehead2021dips,
  title={DIPS-Plus: The Enhanced Database of Interacting Protein Structures for Interface Prediction},
  author={Morehead, Alex and Chen, Chen and Sedova, Ada and Cheng, Jianlin},
  journal={arXiv preprint arXiv:2106.04362},
  year={2021}
}

@article{morehead2021geometric,
  title={Geometric Transformers for Protein Interface Contact Prediction},
  author={Morehead, Alex and Chen, Chen and Cheng, Jianlin},
  journal={arXiv preprint arXiv:2110.02423},
  year={2021}
}

@article{jumper2021highly,
  title={Highly accurate protein structure prediction with AlphaFold},
  author={Jumper, John and Evans, Richard and Pritzel, Alexander and Green, Tim and Figurnov, Michael and Ronneberger, Olaf and Tunyasuvunakool, Kathryn and Bates, Russ and {\v{Z}}{\'\i}dek, Augustin and Potapenko, Anna and others},
  journal={Nature},
  volume={596},
  number={7873},
  pages={583--589},
  year={2021},
  publisher={Nature Publishing Group}
}

@article{hinton2012improving,
  title={Improving neural networks by preventing co-adaptation of feature detectors},
  author={Hinton, Geoffrey E and Srivastava, Nitish and Krizhevsky, Alex and Sutskever, Ilya and Salakhutdinov, Ruslan R},
  journal={arXiv preprint arXiv:1207.0580},
  year={2012}
}

@article{loshchilov2017decoupled,
  title={Decoupled weight decay regularization},
  author={Loshchilov, Ilya and Hutter, Frank},
  journal={arXiv preprint arXiv:1711.05101},
  year={2017}
}

@article{vaswani2017attention,
  title={Attention is all you need},
  author={Vaswani, Ashish and Shazeer, Noam and Parmar, Niki and Uszkoreit, Jakob and Jones, Llion and Gomez, Aidan N and Kaiser, {\L}ukasz and Polosukhin, Illia},
  journal={Advances in neural information processing systems},
  volume={30},
  year={2017}
}

@inproceedings{rao2021msa,
  title={MSA transformer},
  author={Rao, Roshan M and Liu, Jason and Verkuil, Robert and Meier, Joshua and Canny, John and Abbeel, Pieter and Sercu, Tom and Rives, Alexander},
  booktitle={International Conference on Machine Learning},
  pages={8844--8856},
  year={2021},
  organization={PMLR}
}

@article{kotthoff2021dockground,
  title={Dockground scoring benchmarks for protein docking},
  author={Kotthoff, Ian and Kundrotas, Petras J and Vakser, Ilya A},
  journal={Proteins: Structure, Function, and Bioinformatics},
  year={2021},
  publisher={Wiley Online Library}
}

@article{liu2008dockground,
  title={Dockground protein--protein docking decoy set},
  author={Liu, Shiyong and Gao, Ying and Vakser, Ilya A},
  journal={Bioinformatics},
  volume={24},
  number={22},
  pages={2634--2635},
  year={2008},
  publisher={Oxford University Press}
}

@article{tovchigrechko2006gramm,
  title={GRAMM-X public web server for protein--protein docking},
  author={Tovchigrechko, Andrey and Vakser, Ilya A},
  journal={Nucleic acids research},
  volume={34},
  number={suppl\_2},
  pages={W310--W314},
  year={2006},
  publisher={Oxford University Press}
}

@article{pierce2011accelerating,
  title={Accelerating protein docking in ZDOCK using an advanced 3D convolution library},
  author={Pierce, Brian G and Hourai, Yuichiro and Weng, Zhiping},
  journal={PloS one},
  volume={6},
  number={9},
  pages={e24657},
  year={2011},
  publisher={Public Library of Science San Francisco, USA}
}

@inproceedings{gao2021high,
  title={High-Performance Deep Learning Toolbox for Genome-Scale Prediction of Protein Structure and Function},
  author={Gao, Mu and Lund-Andersen, Peik and Morehead, Alex and Mahmud, Sajid and Chen, Chen and Chen, Xiao and Giri, Nabin and Roy, Raj S and Quadir, Farhan and Effler, T Chad and others},
  booktitle={2021 IEEE/ACM Workshop on Machine Learning in High Performance Computing Environments (MLHPC)},
  pages={46--57},
  year={2021},
  organization={IEEE}
}

@article{gao2022proteome,
  title={Proteome-scale Deployment of Protein Structure Prediction Workflows on the Summit Supercomputer},
  author={Gao, Mu and Coletti, Mark and Davidson, Russell B and Prout, Ryan and Abraham, Subil and Hernandez, Benjamin and Sedova, Ada},
  journal={arXiv preprint arXiv:2201.10024},
  year={2022}
}

@article{ulirdita2021fast,
  title={Fast and sensitive taxonomic assignment to metagenomic contigs},
  author={Mirdita, M and Steinegger, M and Breitwieser, F and S{\"o}ding, J and Levy Karin, E},
  journal={Bioinformatics},
  volume={37},
  number={18},
  pages={3029--3031},
  year={2021},
  publisher={Oxford University Press}
}

@article{remmert2012hhblits,
  title={HHblits: lightning-fast iterative protein sequence searching by HMM-HMM alignment},
  author={Remmert, Michael and Biegert, Andreas and Hauser, Andreas and S{\"o}ding, Johannes},
  journal={Nature methods},
  volume={9},
  number={2},
  pages={173--175},
  year={2012},
  publisher={Nature Publishing Group}
}

@article{uniprot2019uniprot,
  title={UniProt: a worldwide hub of protein knowledge},
  author={UniProt Consortium},
  journal={Nucleic acids research},
  volume={47},
  number={D1},
  pages={D506--D515},
  year={2019},
  publisher={Oxford University Press}
}

@article{finn2011hmmer,
  title={HMMER web server: interactive sequence similarity searching},
  author={Finn, Robert D and Clements, Jody and Eddy, Sean R},
  journal={Nucleic acids research},
  volume={39},
  number={suppl\_2},
  pages={W29--W37},
  year={2011},
  publisher={Oxford University Press}
}

@article{bryant2022improved,
  title={Improved prediction of protein-protein interactions using AlphaFold2},
  author={Bryant, Patrick and Pozzati, Gabriele and Elofsson, Arne},
  journal={Nature Communications},
  volume={13},
  number={1},
  pages={1--11},
  year={2022},
  publisher={Nature Publishing Group}
}

@article{c,
  title={Updates to the integrated protein--protein interaction benchmarks: docking benchmark version 5 and affinity benchmark version 2},
  author={Vreven, Thom and Moal, Iain H and Vangone, Anna and Pierce, Brian G and Kastritis, Panagiotis L and Torchala, Mieczyslaw and Chaleil, Raphael and Jim{\'e}nez-Garc{\'\i}a, Brian and Bates, Paul A and Fernandez-Recio, Juan and others},
  journal={Journal of molecular biology},
  volume={427},
  number={19},
  pages={3031--3041},
  year={2015},
  publisher={Elsevier}
}

@article{vreven2015updates,
  title={Updates to the integrated protein--protein interaction benchmarks: docking benchmark version 5 and affinity benchmark version 2},
  author={Vreven, Thom and Moal, Iain H and Vangone, Anna and Pierce, Brian G and Kastritis, Panagiotis L and Torchala, Mieczyslaw and Chaleil, Raphael and Jim{\'e}nez-Garc{\'\i}a, Brian and Bates, Paul A and Fernandez-Recio, Juan and others},
  journal={Journal of molecular biology},
  volume={427},
  number={19},
  pages={3031--3041},
  year={2015},
  publisher={Elsevier}
}

@article{eismann2021hierarchical,
  title={Hierarchical, rotation-equivariant neural networks to select structural models of protein complexes},
  author={Eismann, Stephan and Townshend, Raphael JL and Thomas, Nathaniel and Jagota, Milind and Jing, Bowen and Dror, Ron O},
  journal={Proteins: Structure, Function, and Bioinformatics},
  volume={89},
  number={5},
  pages={493--501},
  year={2021},
  publisher={Wiley Online Library}
}

@article{chen20223d,
  title={3D-equivariant graph neural networks for protein model quality assessment},
  author={Chen, Chen and Chen, Xiao and Morehead, Alex and Wu, Tianqi and Cheng, Jianlin},
  journal={bioRxiv},
  year={2022},
  publisher={Cold Spring Harbor Laboratory}
}

@article{kundrotas2018dockground,
  title={Dockground: a comprehensive data resource for modeling of protein complexes},
  author={Kundrotas, Petras J and Anishchenko, Ivan and Dauzhenka, Taras and Kotthoff, Ian and Mnevets, Daniil and Copeland, Matthew M and Vakser, Ilya A},
  journal={Protein Science},
  volume={27},
  number={1},
  pages={172--181},
  year={2018},
  publisher={Wiley Online Library}
}

@inproceedings{hu2019local,
  title={Local Relation Networks for Image Recognition},
  author={Hu, Han and Zhang, Zheng and Xie, Zhenda and Lin, Stephen},
  booktitle={Proceedings of the IEEE/CVF International Conference on Computer Vision (ICCV)},
  pages={3464--3473},
  year={2019}
}

@article{liu2021swin,
  title={Swin Transformer V2: Scaling Up Capacity and Resolution},
  author={Liu, Ze and Hu, Han and Lin, Yutong and Yao, Zhuliang and Xie, Zhenda and Wei, Yixuan and Ning, Jia and Cao, Yue and Zhang, Zheng and Dong, Li and others},
  journal={arXiv preprint arXiv:2111.09883},
  year={2021}
}

@inproceedings{liu2021swinv2,
  title={Swin Transformer V2: Scaling Up Capacity and Resolution}, 
  author={Ze Liu and Han Hu and Yutong Lin and Zhuliang Yao and Zhenda Xie and Yixuan Wei and Jia Ning and Yue Cao and Zheng Zhang and Li Dong and Furu Wei and Baining Guo},
  booktitle={International Conference on Computer Vision and Pattern Recognition (CVPR)},
  year={2022}
}

@article{xie2022deep,
  title={Deep graph learning of inter-protein contacts},
  author={Xie, Ziwei and Xu, Jinbo},
  journal={Bioinformatics},
  volume={38},
  number={4},
  pages={947--953},
  year={2022},
  publisher={Oxford University Press}
}

@article{fuchs2020se,
  title={Se (3)-transformers: 3d roto-translation equivariant attention networks},
  author={Fuchs, Fabian and Worrall, Daniel and Fischer, Volker and Welling, Max},
  journal={Advances in Neural Information Processing Systems},
  volume={33},
  pages={1970--1981},
  year={2020}
}

@article{kinch2021topology,
  title={Topology evaluation of models for difficult targets in the 14th round of the critical assessment of protein structure prediction (CASP14)},
  author={Kinch, Lisa N and Pei, Jimin and Kryshtafovych, Andriy and Schaeffer, R Dustin and Grishin, Nick V},
  journal={Proteins: Structure, Function, and Bioinformatics},
  volume={89},
  number={12},
  pages={1673--1686},
  year={2021},
  publisher={Wiley Online Library}
}

@article{velivckovic2017graph,
  title={Graph attention networks},
  author={Veli{\v{c}}kovi{\'c}, Petar and Cucurull, Guillem and Casanova, Arantxa and Romero, Adriana and Lio, Pietro and Bengio, Yoshua},
  journal={arXiv preprint arXiv:1710.10903},
  year={2017}
}

@article{ganea2021independent,
  title={Independent SE (3)-Equivariant Models for End-to-End Rigid Protein Docking},
  author={Ganea, Octavian-Eugen and Huang, Xinyuan and Bunne, Charlotte and Bian, Yatao and Barzilay, Regina and Jaakkola, Tommi and Krause, Andreas},
  journal={arXiv preprint arXiv:2111.07786},
  year={2021}
}

@article{stark2022equibind,
  title={Equibind: Geometric deep learning for drug binding structure prediction},
  author={St{\"a}rk, Hannes and Ganea, Octavian-Eugen and Pattanaik, Lagnajit and Barzilay, Regina and Jaakkola, Tommi},
  journal={arXiv preprint arXiv:2202.05146},
  year={2022}
}

@article{baek2021accurate,
  title={Accurate prediction of protein structures and interactions using a three-track neural network},
  author={Baek, Minkyung and DiMaio, Frank and Anishchenko, Ivan and Dauparas, Justas and Ovchinnikov, Sergey and Lee, Gyu Rie and Wang, Jue and Cong, Qian and Kinch, Lisa N and Schaeffer, R Dustin and others},
  journal={Science},
  volume={373},
  number={6557},
  pages={871--876},
  year={2021},
  publisher={American Association for the Advancement of Science}
}

@inproceedings{chen2020deep,
  title={Deep ranking in template-free protein structure prediction},
  author={Chen, Xiao and Akhter, Nasrin and Guo, Zhiye and Wu, Tianqi and Hou, Jie and Shehu, Amarda and Cheng, Jianlin},
  booktitle={Proceedings of the 11th ACM International Conference on Bioinformatics, Computational Biology and Health Informatics},
  pages={1--10},
  year={2020}
}

@article{chen2022distema,
  title={DISTEMA: distance map-based estimation of single protein model accuracy with attentive 2D convolutional neural network},
  author={Chen, Xiao and Cheng, Jianlin},
  journal={BMC bioinformatics},
  volume={23},
  number={3},
  pages={1--14},
  year={2022},
  publisher={Springer}
}

@article{wu2021deepdist,
  title={DeepDist: real-value inter-residue distance prediction with deep residual convolutional network},
  author={Wu, Tianqi and Guo, Zhiye and Hou, Jie and Cheng, Jianlin},
  journal={BMC bioinformatics},
  volume={22},
  number={1},
  pages={1--17},
  year={2021},
  publisher={Springer}
}

@article{xu2015empirical,
  title={Empirical evaluation of rectified activations in convolutional network},
  author={Xu, Bing and Wang, Naiyan and Chen, Tianqi and Li, Mu},
  journal={arXiv preprint arXiv:1505.00853},
  year={2015}
}

@article{berman2000protein,
  title={The protein data bank},
  author={Berman, Helen M and Westbrook, John and Feng, Zukang and Gilliland, Gary and Bhat, Talapady N and Weissig, Helge and Shindyalov, Ilya N and Bourne, Philip E},
  journal={Nucleic acids research},
  volume={28},
  number={1},
  pages={235--242},
  year={2000},
  publisher={Oxford University Press}
}

@article{jendrusch2021alphadesign,
  title={AlphaDesign: A de novo protein design framework based on AlphaFold},
  author={Jendrusch, Michael and Korbel, Jan O and Sadiq, S Kashif},
  journal={bioRxiv},
  year={2021},
  publisher={Cold Spring Harbor Laboratory}
}
}

%%%%%%%%%%%%%%%%%%%%%%%%%%%%%%%%%%%%%%%%%%%%%%%%%%%%%%%%%%%%
% \section*{Checklist}
% \input{sections/Checklist}

\appendix

\section{Additional Results}
\label{sec:appendix_a}
Table \ref{tab: HAF2_seed_res} shows models' hit rate performance on the HAF2 dataset, while Table \ref{tab: HAF2_seed_loss} displays models' ranking loss performance on the HAF2 dataset. Likewise, Table \ref{tab: DBM55_seed_res} displays models' hit rate performance on the DBM55-AF2 dataset, while Table \ref{tab: DBM55_seed_loss} shows their ranking loss performance on the DBM55-AF2 dataset. Best performances are highlighted in \textbf{bold}. We note for readers that \textsc{DProQ}\_222 is the model for which we reported our final results in the main text.

\begin{table}[H]
\caption{HAF2 dataset hit rate of GNN\_DOVE and \textsc{DProQ} with five different random seeds. The number proceeding \_ is a specific random seed.}
\label{tab: HAF2_seed_res}
\centering
\begin{tabular}{lllllll}
\toprule
Target  & \textsc{DProQ}\_111     & \textsc{DProQ}\_222      & \textsc{DProQ}\_520      & \textsc{DProQ}\_888      & \textsc{DProQ}\_999     & GNN\_DOVE     \\ \midrule
7AOH    & 10/10/10 & 10/10/10 & 10/10/10 & 10/10/1  & 10/10/1  & 9/9/0     \\
7D7F    & 0/0/0    & 0/0/0    & 0/0/0    & 0/0/0    & 0/0/0    & 0/0/0     \\
7AMV    & 10/10/10 & 10/10/10 & 10/10/10 & 10/10/1  & 10/10/1  & 10/10/6   \\
7OEL    & 10/10/0  & 10/10/0  & 10/10/0  & 10/10/0  & 10/10/0  & 10/10/0   \\
7O28    & 10/10/0  & 10/10/0  & 10/10/0  & 10/10/0  & 10/10/0  & 10/10/0   \\
7ALA    & 0/0/0    & 0/0/0    & 0/0/0    & 0/0/0    & 0/0/0    & 0/0/0     \\
7MRW    & 0/0/0    & 5/4/0    & 0/0/0    & 0/0/0    & 2/2/0    & 0/0/0     \\
7OZN    & 0/0/0    & 0/0/0    & 0/0/0    & 0/0/0    & 0/0/0    & 0/0/0     \\
7D3Y    & 6/0/0    & 2/0/0    & 7/0/0    & 4/0/0    & 8/0/0    & 0/0/0     \\
7NKZ    & 10/10/2  & 10/10/2  & 10/10/1  & 10/10/1  & 10/10/2  & 10/9/9    \\
7LXT    & 0/0/0    & 1/1/0    & 0/0/0    & 0/0/0    & 0/0/0    & 1/0/0     \\
7KBR    & 10/10/ & 10/10/10 & 10/10/10 & 10/10/9  & 10/10/9  & 10/10/9   \\
7O27    & 10/10/0  & 10/10/0  & 10/10/0  & 10/10/0  & 10/10/0  & 10/4/0    \\ \midrule
Summary & 8/7/4   & \textbf{10/9/4}   & 8/7/4    & 8/7/4    & 9/8/4    & 8/7/3     \\ \bottomrule
\end{tabular}
\end{table}

\begin{table}[H]
\caption{HAF2 dataset ranking loss of GNN\_DOVE and \textsc{DProQ} with five different random seeds. The number proceeding \_ is a specific random seed.}
\label{tab: HAF2_seed_loss}
\centering
\begin{tabular}{lllllll}
\toprule
Target  & \textsc{DProQ}\_111    & \textsc{DProQ}\_222    & \textsc{DProQ}\_520    & \textsc{DProQ}\_888    & \textsc{DProQ}\_999   & GNN\_DOVE     \\ \midrule
7AOH    & 0.05          & 0.066         & 0.026         & 0.026         & 0.066        & 0.928         \\
7D7F    & 0.471         & 0.471         & 0.47          & 0.47          & 0.471        & 0.003         \\
7AMV    & 0.019         & 0.01          & 0.019         & 0.019         & 0.017        & 0.342         \\
7OEL    & 0.135         & 0.062         & 0.063         & 0.135         & 0.063        & 0.21          \\
7O28    & 0.021         & 0.029         & 0.151         & 0.021         & 0.02         & 0.244         \\
7ALA    & 0.226         & 0.232         & 0.234         & 0.227         & 0.234        & 0.226         \\
7MRW    & 0.555         & 0.085         & 0.599         & 0.555         & 0.555        & 0.598         \\
7OZN    & 0.412         & 0.409         & 0.493         & 0.412         & 0.409        & 0.457         \\
7D3Y    & 0.326         & 0.326         & 0.326         & 0.326         & 0.326        & 0.295         \\
7NKZ    & 0.164         & 0.164         & 0.175         & 0.164         & 0.164        & 0.459         \\
7LXT    & 0.586         & 0.586         & 0.586         & 0.586         & 0.586        & 0.295         \\
7KBR    & 0.193         & 0.068         & 0.095         & 0.026         & 0.152        & 0.068         \\
7O27    & 0.03          & 0.03          & 0.079         & 0.079         & 0.03         & 0.334         \\ \midrule
Mean    & 0.245 ± 0.197 & \textbf{0.195 ± 0.185} & 0.255 ± 0.207 & 0.234 ± 0.204 & 0.238 ± 0.201 & 0.343 ± 0.228 \\ \bottomrule
\end{tabular}
\end{table}

\begin{table}[H]
\caption{DBM55-AF2 dataset hit rate of GNN\_DOVE and \textsc{DProQ} with five different random seeds. The number proceeding \_ is the specific random seed.}
\label{tab: DBM55_seed_res}
\centering
\begin{tabular}{lllllll}
\toprule
Target  & \textsc{DProQ}\_111         & \textsc{DProQ}\_222      & \textsc{DProQ}\_520      & \textsc{DProQ}\_888      & \textsc{DProQ}\_999     & GNN\_DOVE     \\ \midrule
6AL0    & 10/2/0             & 9/2/0           & 10/2/0          & 10/2/0          & 10/1/0         & 6/0/0     \\
3SE8    & 10/10/0            & 8/8/0           & 8/8/0           & 10/10/0         & 8/8/0          & 3/0/0     \\
5GRJ    & 10/10/0            & 10/10/0         & 9/9/0           & 9/9/0           & 9/9/0          & 3/2/0     \\
6A77    & 7/7/0              & 7/7/0           & 8/8/0           & 7/7/0           & 8/8/0          & 0/0/0     \\
4M5Z    & 10/10/1            & 10/10/1         & 10/10/0         & 10/10/1         & 10/10/0        & 10/10/0   \\
4ETQ    & 1/1/0              & 1/1/0           & 1/1/0           & 1/1/0           & 1/1/0          & 0/0/0     \\
5CBA    & 10/10/0            & 10/10/1         & 10/10/1         & 10/10/1         & 10/10/1        & 10/10/3   \\
5WK3    & 0/0/0              & 0/0/0           & 0/0/0           & 0/0/0           & 0/0/0          & 1/0/0     \\
5Y9J    & 6/0/0              & 4/0/0           & 8/0/0           & 5/0/0           & 4/0/0          & 0/0/0     \\
6B0S    & 10/10/0            & 10/10/0         & 10/10/0         & 10/10/0         & 10/10/0        & 10/10/0   \\
5HGG    & 8/0/0              & 8/0/0           & 8/0/0           & 8/0/0           & 8/0/0          & 8/0/0     \\
6A0Z    & 0/0/0              & 0/0/0           & 0/0/0           & 0/0/0           & 0/0/0          & 2/0/0     \\
3U7Y    & 2/2/1              & 2/2/1           & 2/2/1           & 2/2/1           & 2/2/1          & 2/2/1     \\
3WD5    & 9/8/0              & 10/8/0          & 8/8/0           & 9/8/0           & 9/8/0          & 0/0/0     \\
5KOV    & 0/0/0              & 0/0/0           & 0/0/0           & 0/0/0           & 0/0/0          & 1/0/0     \\ \midrule
Summary & 12/10/2            & \textbf{12/10/3} & 12/10/2  & \textbf{12/10/3}      & 12/10/2        & 10/4/1    \\ \bottomrule
\end{tabular}
\end{table}

\begin{table}[H]
\caption{DBM55-AF2 dataset ranking loss of GNN\_DOVE and \textsc{DProQ} with five different random seeds. The number proceeding \_ is the specific random seed.}
\label{tab: DBM55_seed_loss}
\centering
\begin{tabular}{lllllll}
\toprule
Target  & \textsc{DProQ}\_111         & \textsc{DProQ}\_222      & \textsc{DProQ}\_520      & \textsc{DProQ}\_888      & \textsc{DProQ}\_999     & GNN\_DOVE     \\ \midrule
6AL0    & 0.156         & 0.0           & 0.156         & 0.156         & 0.156        & 0.424         \\
3SE8    & 0.041         & 0.079         & 0.041         & 0.041         & 0.068        & 0.735         \\
5GRJ    & 0.095         & 0.024         & 0.012         & 0.012         & 0.124        & 0.776         \\
6A77    & 0.062         & 0.037         & 0.0           & 0.037         & 0.062        & 0.591         \\
4M5Z    & 0.242         & 0.015         & 0.015         & 0.015         & 0.251        & 0.221         \\
4ETQ    & 0.0           & 0.0           & 0.75          & 0.748         & 0.75         & 0.759         \\
5CBA    & 0.038         & 0.052         & 0.054         & 0.052         & 0.071        & 0.019         \\
5WK3    & 0.123         & 0.114         & 0.186         & 0.114         & 0.186        & 0.087         \\
5Y9J    & 0.0           & 0.0           & 0.0           & 0.0           & 0.0          & 0.382         \\
6B0S    & 0.0           & 0.081         & 0.0           & 0.0           & 0.081        & 0.081         \\
5HGG    & 0.007         & 0.051         & 0.051         & 0.051         & 0.051        & 0.121         \\
6A0Z    & 0.207         & 0.207         & 0.214         & 0.207         & 0.207        & 0.062         \\
3U7Y    & 0.0           & 0.0           & 0.0           & 0.0           & 0.021        & 0.756         \\
3WD5    & 0.109         & 0.011         & 0.0           & 0.192         & 0.109        & 0.672         \\ 
5KOV    & 0.087         & 0.065         & 0.087         & 0.087         & 0.09         & 0.0           \\ \midrule
Mean    & 0.078 ± 0.076 & \textbf{0.049 ± 0.054} & 0.104 ± 0.186 & 0.114 ± 0.182 & 0.148 ± 0.174 & 0.379 ± 0.298 \\ \bottomrule
\end{tabular}
\end{table}

\section{Additional Dataset Materials}
\label{sec:appendix_b}
Tables \ref{tab:maf2}, \ref{tab:haf2}, and \ref{tab:DBMAF2} proceed to list dataset details for the MFA2, HAF2, and DBM55-AF2 datasets, respectively, while Table \ref{tab:graph_features} describes the features encoded in \textsc{DProQ}'s input graphs.

\textbf{Labels.} MAF2, HAF2, and DBM55-AF2's DockQ scores were produced using the DockQ software available at \href{https://github.com/bjornwallner/DockQ/}{\texttt{https://github.com/bjornwallner/DockQ/}}. The Docking Decoy dataset provides interface root mean squared deviations (iRMSDs), ligand RMSDs (LRMSs), and fractions of native contacts for each decoy ($f_{nat}$). We use Equations \ref{eq:dockq_rmsd} and \ref{eq:dockq_main} with $d_{1}=8.5$ and $d_{2}=1.5$ to generate the Docking Decoy dataset's labels. To derive classification labels, based on Figure 1 in \cite{basu2016dockq}, we convert each decoy's DockQ label to $0$, $1$, $2$, $3$ to represent Incorrect, Acceptable, Medium, and High qualities, respectively.

\begin{equation}
\label{eq:dockq_rmsd}
    RMSD_{\text {scaled }}\left(RMSD, d_{i}\right)=\frac{1}{1+\left(\frac{RMSD}{d_{i}}\right)^{2}}
\end{equation}

\begin{equation}
\label{eq:dockq_main}
    \text{DockQ}=\frac{1}{3} \left(F_{\text {nat }}+ RMSD_{\text {scaled }}\left(LRMSD, d_{1}\right) + RMSD_{\text {scaled }}\left(iRMSD, d_{2}\right)\right).
\end{equation}

% MAF2 dataset
\textbf{Multimer-AF2 dataset.} Drawing loose inspiration from \cite{morehead2021dips}, for this work, we assembled the new Multimer-AF2 (MAF2) dataset comprised of multimeric structures predicted by AlphaFold 2 \cite{jumper2021highly} and AlphaFold-Multimer \cite{evans2021protein}, using the latest structure prediction pipeline for the Summit supercomputer \cite{gao2021high, gao2022proteome}. Originating from the EVCoupling \cite{hopf2019evcouplings} and DeepHomo \cite{yan2021accurate} datasets, the proteins for which we predicted structures using a combination of both AlphaFold methods consist of heteromers and homomers, respectively.\par

For structures we predicted using AlphaFold2, before structure prediction, we inserted 20-glycine residues in between chains in the multimeric input sequences to facilitate multimeric structure generation with monomeric input sequences. For structures predicted with AlphaFold-Multimer, we directly input the original multimeric sequences into the model's prediction pipeline. Unlike AlphaFold2 and Alphafold-Multimer default settings, we only maintain the TOP1 decoy for modeling. In designing a cross-validation partitioning scheme for the MAF2 dataset, we first randomly split predicted proteins into an $70/20/10\%$ training, validation, and testing split. After making this split, we then removed from the validation split any proteins with at least $30\%$ sequence identity w.r.t the test split's proteins. Similarly, we then filtered from the training split proteins with at least $30\%$ sequence identity w.r.t either the validation or test split.\par

\begin{table}[H]
\centering
\caption{Summary of the MAF2 dataset.}
\label{tab:maf2}
\begin{tabular}{lllllll}
\toprule
             & Incorrect & Acceptable & Medium & High & Mean DockQ  & Median DockQ \\
\midrule
AlphaFold-Multimer & 590       & 371        & 1237   & 1878 & 0.66         & 0.78          \\
AlphaFold 2          & 1301      & 956        & 301    & 1380 & 0.52         & 0.67          \\
\midrule
Total      & 1891      & 1327       & 2775   & 3258 & 0.59         & 0.74          \\
\bottomrule
\end{tabular}
\end{table}

% HAF2 dataset
\textbf{Heterodimer-AF2 dataset.} For our Heterodimer-AF2 (HAF2) dataset, we collected the structures of heterodimers from Protein Data Bank (PDB) with release dates between September 2021 and November 2021. After extracting their corresponding single-chain protein structures, we used MMseqs2 \cite{ulirdita2021fast} to reduce their redundancy using a 40\% sequence identity threshold. We considered a pair of monomers as heterodimers if they share the same PDB code and according to whether their minimum heavy atom distance was larger than 6 \si{\angstrom}. Finally, we only selected heterodimers from different complexes and further randomly-sampled 70 heterodimers for use as test complexes.\par

For each heterodimer, we used a set of variants of AlphaFold-Multimer \cite{evans2021protein} derived from the original software to generate different models. Two main changes were made to the original AlphaFold-Multimer system. Firstly, we used HHblits \cite{remmert2012hhblits} and JackHMMER \cite{finn2011hmmer} to search the sequence of each chain against the UniRef30, UniRef90 \cite{uniprot2019uniprot}, and UniProt \cite{uniprot2019uniprot} databases. Then, we applied different complex alignment concatenation strategies, including extracting interactions between chains in the complex using mappings from UniProt IDs to PDB codes in the PDB database; species information extracted using different approaches \cite{hopf2019evcouplings, bryant2022improved}; genomics distance calculated from UniProt IDs \cite{uniprot2019uniprot}; and interaction information derived from the STRING database. Secondly, we searched the sequence of each chain against different template databases, including PDB70 and our in-house template database built from the PDB. Finally, we paired the templates based on their PDB codes. The paired templates are used to generate template features for AlphaFold-Multimer to generate structures.\par

\begin{table}[H]
\centering
\caption{Summary of the HAF2 dataset.}
\label{tab:haf2}
\begin{tabular}{lllllll}
\toprule
Target & Incorrect & Acceptable & Medium & High & Mean DockQ & Median DockQ     \\ \midrule
7AOH &1 &1 &24 &64 &0.83 &0.89 \\
7D7F &110 &5 &0 &0 &0.03 &0.007 \\
7AMV &0 &0 &5 &65 &0.84 &0.869 \\
7OEL &0 &1 &114 &0 &0.71 &0.722 \\
7O28 &0 &0 &115 &0 &0.72 &0.734 \\
7ALA &99 &1 &0 &0 &0.03 &0.017 \\
7MRW &40 &4 &46 &0 &0.32 &0.492 \\
7OZN &96 &12 &2 &0 &0.07 &0.015 \\
7D3Y &70 &35 &0 &0 &0.13 &0.035 \\
7NKZ &0 &1 &16 &98 &0.87 &0.915 \\
7LXT &41 &1 &73 &0 &0.4 &0.564 \\
7KBR &0 &0 &1 &114 &0.91 &0.922 \\
7O27 &0 &23 &92 &0 &0.65 &0.705 \\
\midrule
Total &457 &84 &488 &341 &0.5 &0.63 \\ \bottomrule
\end{tabular}
\end{table}

% DBM55-AF2
\textbf{Docking Benchmark 5.5-AF2 dataset.} The Docking Benchmark 5.5-AF2 (DBM55-AF2) dataset is our final blind test dataset. It is comprised of heteromers derived from the Docking Benchmark 5.5 dataset \cite{vreven2015updates}. Specifically, this test dataset consists of five randomly-sampled heteromers from each of the Benchmark 5.5 dataset's three difficulty categories (i.e., Rigid-Body, Medium, Difficult). After obtaining our targets for this test dataset, we calculated the DockQ score for each decoy and then filtered out the target if it does not contain any decoy of Acceptable or higher-quality. Moreover, we performed 30\% sequence identity filtering w.r.t our training data for this test dataset. Table \ref{tab:DBMAF2} shows each targets' decoy accuracy level details.\par

\begin{table}[H]
\centering
\caption{Summary of the Docking Benchmark 5.5-AF2 dataset.}
\label{tab:DBMAF2}
\begin{tabular}{lllllll}
\toprule
Target & Incorrect & Acceptable & Medium & High & Mean DockQ & Median DockQ \\ \midrule
6AL0   & 4         & 24         & 2      & 0    & 0.28        & 0.26          \\
3SE8   & 15        & 3          & 12     & 0    & 0.33        & 0.19          \\
5GRJ   & 8         & 1          & 21     & 0    & 0.52        & 0.68          \\
6A77   & 22        & 0          & 8      & 0    & 0.23        & 0.08          \\
4M5Z   & 0         & 0          & 29     & 1    & 0.61        & 0.59          \\
4ETQ   & 29        & 0          & 1      & 0    & 0.08        & 0.04          \\
5CBA   & 0         & 0          & 24     & 6    & 0.78        & 0.78          \\
5WK3   & 27        & 3          & 0      & 0    & 0.17        & 0.18          \\
5Y9J   & 22        & 8          & 0      & 0    & 0.16        & 0.12          \\
6B0S   & 0         & 0          & 30     & 0    & 0.57        & 0.57          \\
5HGG   & 2         & 27         & 0      & 0    & 0.29        & 0.29          \\
6A0Z   & 27        & 3          & 0      & 0    & 0.1         & 0.08          \\
3U7Y   & 28        & 0          & I      & 1    & 0.1         & 0.04          \\
3WD5   & 16        & 4          & 10     & 0    & 0.29        & 0.18          \\
5KOV   & 28        & 2          & 0      & 0    & 0.19        & 0.19          \\ \midrule
Total  & 228       & 75         & 138    & 8    & 0.32        & 0.19          \\ \bottomrule
\end{tabular}
\end{table}

\textbf{Featurization.} $35$ node features and $6$ edge features are assigned to each graph $\mathcal{G}$. Table \ref{tab:graph_features} summarizes each feature's details. For each graph $\mathcal{G}$, the node features' shape is $N \times 35$, and the edge features' shape is $E \times 6$, where $N$ is the graph's node count and $E$ is its edge count.

For node features, we first produced one-hot encodings based on a given protein complex's residue types. Then, we used DSSP 3.0.0 \cite{joosten2010series, kabsch1983dictionary} and BioPython 1.79 \cite{cock2009biopython} to calculate relative accessible surface areas as well as $\phi$ and $\psi$ angles for each residue. We then conducted a min-max normalization of the $\phi$ and $\psi$ angle values to scale their value ranges from [$-180$, $180$] to [$0$, $1$]. We also added a Graph Laplacian positional encoding \cite{dwivedi2020generalization} to each node. 

For the edge features in \textsc{DProQ}'s graphs, we calculated the alpha carbon-alpha carbon (C$\alpha$-C$\alpha$), beta carbon-beta carbon (C$\beta$-C$\beta$), and nitrogen-oxygen (N-O) distances to serve as edge features. As another edge feature, we consider two residues to be in contact with one another if their C$\alpha$-C$\alpha$ distance is less than $8$ \si{\angstrom}, yielding a binary edge feature with a value of $1$ for residues in contact. To encode chain information into our graphs' edges, we introduce a new permutation-invariant chain encoding. In particular, for an edge's vertices, if they are adjacent to one another w.r.t. the protein complex's amino acid sequence and they belong to the same chain, we encoded a feature value of $1$ for this edge and $0$ otherwise. Lastly, we include an edge-wise positional encoding \cite{morehead2021geometric} for each edge. 

\begin{table}[H]
\caption{Summary of \textsc{DProQ}'s node and edge features. Here, $N$ and $E$ denote the number of nodes and edges in $\mathcal{G}$, respectively.}
\label{tab:graph_features}
\centering
\begin{tabular}{clll}
\toprule
                                             & \multicolumn{1}{c}{Feature} & \multicolumn{1}{c}{Type} & Shape \\ \midrule
\multirow{1}{*}{Node Features}                 & One-hot encoding of residue type          & Categorical                        & $N \times 21$          \\
                                             & Three types of secondary structure    & Categorical                        & $N \times 3 $          \\  
                                             & Relative accessible surface area      & Numeric                          & $N \times 1 $          \\  
                                             & $\phi$ angle                       & Numeric                          & $N \times 1 $          \\  
                                             & $\psi$ angle                          & Numeric                          & $N \times 1 $          \\  
                                             & Graph Laplacian positional encoding   & Numeric                          & $N \times 8 $          \\ \midrule
\multirow{1}{*}{Edge Features}                 & C$\alpha$-C$\alpha$ distance          & Numeric                          & $E \times 1 $          \\ 
                                             & C$\beta$-C$\beta$ distance            & Numeric                          & $E \times 1 $          \\  
                                             & N-O distance                          & Numeric                          & $E \times 1 $          \\  
                                             & Inter-chain contact encoding          & Categorical                        & $E \times 1 $          \\  
                                             & Permutation-invariant chain encoding  & Categorical                        & $E \times 1 $          \\  
                                             & Edge positional encoding              & Numeric                          & $E \times 1 $          \\ \midrule
\multicolumn{1}{l}{\multirow{1}{*}{Total}}     & Node features                         &                                    & $N \times 35$          \\  
\multicolumn{1}{l}{}                         & Edge features                         &                                    & $E \times 6 $          \\ \bottomrule
\end{tabular}
\end{table}

\section{Implementation Details}
\label{sec:appendix_c}
\textbf{Hardware Used.} \textsc{DProQ} was trained using two Nvidia GeForce RTX 2080 Super GPUs in a data-parallel manner. With a batch size of $16$ and a gradient accumulation batch size of $32$, its effective batch size during training was $512$.\par

\textbf{Software Used.} We implemented \textsc{DProQ} using PyTorch \cite{NEURIPS2019_9015}, PyTorch Lightning \cite{Falcon_PyTorch_Lightning_2019}, and the Deep Graph Library \cite{wang2019dgl}. PyTorch Lightning was used to facilitate model checkpointing, metrics reporting, and distributed data parallelism across 2 RTX 2080 Super GPUs. A more in-depth description of the software environment and data used to train and run inference with our models can be found at \href{https://github.com/BioinfoMachineLearning/DProQ}{\texttt{https://github.com/BioinfoMachineLearning/DProQ}}.\par

\textbf{Further hyperparameters.} \textsc{DProQ} models are optimized using the AdamW optimizer \cite{loshchilov2017decoupled} with $\beta_{1}=0.9$, $\beta_{2}=0.999$, and a weight decay rate of $0.002$. Each model uses an initial learning rate of $50^{-3}$, where this initial learning rate is then halved every $16$ epochs. During model training, we employ an early stopping patience period of $15$ epochs to prevent our models from overfitting. Models with the lowest loss $\mathcal{L}$ on our validation split are then tested on all our sequence-filtered test datasets.\par

\begin{table}[H]
\label{tab:par}
\centering
\caption{Hyperparameter search space for all \textsc{DProQ} models through which we searched to obtain strong performance on our validation split. The final parameters for the standard \textsc{DProQ} model are in \textbf{bold}.}
\begin{tabular}{ll}
\toprule
Hyperparameter   & Search Space                             \\
\midrule
Weight of $\mathcal{L}_{C}$ ($w_{\mathcal{L}_{C}}$)     & 0.1 (Based on Loss on Validation Split)      \\
Weight of $\mathcal{L}_{R}$ ($w_{\mathcal{L}_{R}}$)     & \textbf{0.9},  0.8,  0.7,  0.6, 0.5      \\
Number of \textsc{GGT} Layers    & 1, \textbf{2}, 3, 4, 6 \\
\textsc{GGT} Dropout Rate      & 0.1, 0.2, 0.3, \textbf{0.4}, 0.5         \\
Read-Out Module Dropout Rate & 0.1, 0.2, 0.3, 0.4, \textbf{0.5}         \\
Number of Attention Heads     & 8 (Based on Loss on Validation Split) \\
Hidden Dimension    & 32, \textbf{64}, 128 \\
Non-Linearities     & LeakyReLU (Based on Loss on Validation Split) \\
Learning Rate    & 0.001, \textbf{0.005}, 0.01, 0.05, 0.1 \\
Weight Decay Rate     & 0.001, \textbf{0.002}, 0.01, 0.02        \\
Normalization    & LayerNorm, \textbf{BatchNorm}            \\
Graph Pooling Operator      & Mean, Max, \textbf{Sum}                  \\
\bottomrule
\end{tabular}
\end{table}

\end{document}